\documentclass[twoside,twocolumn,9pt]{article}
\usepackage{extsizes}
\usepackage[super,sort&compress,comma]{natbib} 
\usepackage[version=3]{mhchem}
\usepackage[left=1.5cm, right=1.5cm, top=1.785cm, bottom=2.0cm]{geometry}
\usepackage{balance}
\usepackage{mathptmx}
\usepackage{sectsty}
\usepackage{graphicx} 
\usepackage{booktabs} 
\usepackage{amsmath}
\usepackage{amssymb}
\usepackage{bm}
\usepackage{lastpage}
\usepackage[format=plain,justification=justified,singlelinecheck=false,font={stretch=1.125,small,sf},labelfont=bf,labelsep=space]{caption}
\usepackage{float}
\usepackage{placeins}
\usepackage{fancyhdr}
\usepackage{fnpos}
\usepackage[english]{babel}
\addto{\captionsenglish}{%
  
}
\usepackage{array}
\usepackage{droidsans}
\usepackage{charter}
\usepackage[T1]{fontenc}
\usepackage{lmodern}
\usepackage[dvipsnames]{xcolor}
\usepackage{setspace}
\usepackage[compact]{titlesec}
\usepackage{xr-hyper}
\usepackage[hidelinks]{hyperref}
\hypersetup{
  pdftitle={A budget-dependent crossover between coverage- and response-based training-set selection for machine-learned interatomic potentials},
  pdfauthor={Jia Bi and Alin-Marin Elena}
}

\usepackage{epstopdf}

\definecolor{cream}{RGB}{222,217,201}
\definecolor{ddblue}{RGB}{148,166,184}
\newcolumntype{P}[1]{>{\raggedright\arraybackslash}p{#1}}

\begin{document}

\pagestyle{fancy}
\thispagestyle{plain}
\fancypagestyle{plain}{
\renewcommand{\headrulewidth}{0pt}
}

\makeFNbottom
\makeatletter
\renewcommand\LARGE{\@setfontsize\LARGE{15pt}{17}}
\renewcommand\Large{\@setfontsize\Large{12pt}{14}}
\renewcommand\large{\@setfontsize\large{10pt}{12}}
\renewcommand\footnotesize{\@setfontsize\footnotesize{7pt}{10}}
\makeatother

\renewcommand{\thefootnote}{\fnsymbol{footnote}}
\renewcommand\footnoterule{\vspace*{1pt}%
\color{cream}\hrule width 3.5in height 0.4pt \color{black}\vspace*{5pt}} 
\setcounter{secnumdepth}{5}

\makeatletter 
\renewcommand\@biblabel[1]{#1}            
\renewcommand\@makefntext[1]%
{\noindent\makebox[0pt][r]{\@thefnmark\,}#1}
\makeatother 
\renewcommand{\figurename}{\small{Fig.}~}
\sectionfont{\sffamily\Large}
\subsectionfont{\normalsize}
\subsubsectionfont{\bf}
\setstretch{1.125} 
\setlength{\skip\footins}{0.8cm}
\setlength{\footnotesep}{0.25cm}
\setlength{\jot}{10pt}
\titlespacing*{\section}{0pt}{4pt}{4pt}
\titlespacing*{\subsection}{0pt}{15pt}{1pt}

\fancyfoot{}
\fancyfoot[CO]{\footnotesize\sffamily Digital Discovery, [year], [vol.],}
\fancyfoot[CE]{\footnotesize\sffamily Digital Discovery, [year], [vol.],}
\fancyfoot[RO]{\footnotesize{\sffamily{1--\pageref{LastPage} ~\textbar  \hspace{2pt}\thepage}}}
\fancyfoot[LE]{\footnotesize{\sffamily{\thepage~\textbar\hspace{3.45cm} 1--\pageref{LastPage}}}}
\fancyhead{}
\renewcommand{\headrulewidth}{0pt} 
\renewcommand{\footrulewidth}{0pt}
\setlength{\arrayrulewidth}{1pt}
\setlength{\columnsep}{6.5mm}
\setlength\bibsep{1pt}

\makeatletter 
\newlength{\figrulesep} 
\setlength{\figrulesep}{0.5\textfloatsep} 

\newcommand{\topfigrule}{\vskip-1pt{\color{cream}\hrule height 1.5pt}\vskip\figrulesep}

\newcommand{\botfigrule}{\vskip\figrulesep{\color{cream}\hrule height 1.5pt}\vskip-2pt}

\newcommand{\dblfigrule}{\vskip-1pt{\color{cream}\hrule height 1.5pt}\vskip\figrulesep}

\makeatother

\twocolumn[
  \begin{@twocolumnfalse}
 {\raisebox{8pt}{\sffamily\bfseries\fontsize{18}{20}\selectfont Digital Discovery}\hfill\raisebox{0pt}[0pt][0pt]{\includegraphics[height=55pt]{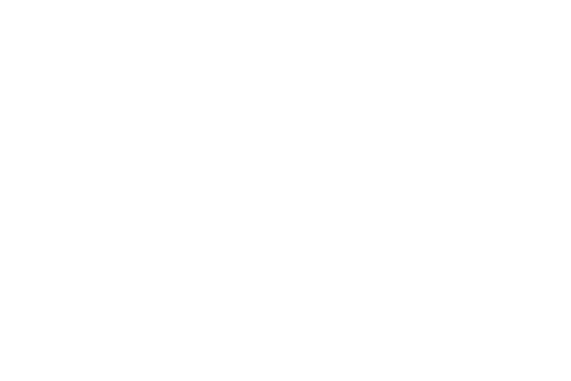}}\\[1ex]
{\setlength{\fboxsep}{0pt}\colorbox{ddblue}{\parbox[c][18pt][c]{18.5cm}{\hspace{0.45cm}\color{white}\sffamily\bfseries\fontsize{8}{9}\selectfont PAPER}}}}\par
\vspace{1em}
\sffamily
\begin{tabular}{@{}m{4.5cm} p{13.5cm}@{}}

\includegraphics{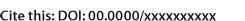} & \noindent\LARGE{\textbf{A budget-dependent crossover between coverage- and response-based training-set selection for machine-learned interatomic potentials$^\dag$}} \\
\vspace{0.3cm} & \vspace{0.3cm} \\

 & \noindent\large{Jia Bi,$^{\ast}$\textit{$^{a}$} Alin-Marin Elena,$^{\ast}$\textit{$^{b}$}} \\

\includegraphics{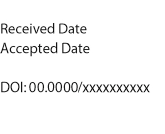} & \noindent\normalsize{Selecting compact training sets for machine-learned interatomic potentials requires deciding whether to preserve structural diversity or target configurations on which models disagree. The better choice can depend on how much data is retained, making a comparison at one training-set size insufficient. Here we link selection criteria to prediction accuracy through a budget-resolved comparison of retrained MACE models on GAP-20 Carbon and pooled revised MD17. Structural coverage is compared with a response-guided selector that targets disagreement between a coverage-trained model and a full-data reference. This retrospective response witness tests the value of model disagreement for compressing an already labelled pool. At 5\%, coverage gives smaller absolute deviations from the full-data error than random sampling across four force endpoints in both datasets. The witness has larger deviations than coverage at 1\% and 5\%, but the ordering reverses at 20\%. At 20\%, witness-selected models also lower direct held-out force errors by 0.46--5.89\% relative to coverage, with all eight paired training-seed intervals favouring the witness. Six errors fall below the full-data reference. Mean force-error reductions are 0.164--0.167~meV~$\text{\AA}^{-1}$, with larger gains for tail and masked endpoints. Complementary analyses show that learned similarity preserves the coverage ranking, while selecting by frozen-model error gives higher error than embedding coverage. These findings establish retained-data budget as a deciding variable in atomistic training-set selection and provide a direct test of when response-guided compression improves on structural coverage.} \\

\end{tabular}

 \end{@twocolumnfalse} \vspace{0.6cm}

  ]

\renewcommand*\rmdefault{bch}\normalfont\upshape
\rmfamily
\section*{}
\vspace{-1cm}

\footnotetext{\textit{$^{a}$~Science and Technology Facilities Council, Harwell Campus, Didcot, United Kingdom. E-mail: Jia.Bi@stfc.ac.uk}}
\footnotetext{$^{\ast}$~Corresponding author. \textit{$^{b}$~Science and Technology Facilities Council, Keckwick Lane, Daresbury, United Kingdom. E-mail: alin-marin.elena@stfc.ac.uk}}


\footnotetext{\dag~Supplementary Information available: supplementary methods, extended figures, tables and source-data descriptions.}


\section{Introduction}
\label{sec:intro}

Machine-learned interatomic potentials (MLIPs) make molecular and materials simulations accessible beyond the cost of repeated electronic-structure calculations. Neural-network, kernel and systematically improvable potentials learn energies and forces from representations of local atomic environments~\cite{behlerparrinello,gap,snap,mtp,ace,soap}. Message-passing and equivariant graph architectures have further improved data efficiency and force prediction~\cite{schnet,painn,nequip,allegro,mace}. Their accuracy still depends on which configurations enter training. When an existing labelled dataset is compressed, the task is to retain the configurations needed for accurate predictions while discarding redundant examples. At a fixed training-set size, should selection favour a broad range of structures or the configurations on which a model struggles most?

Growing atomistic datasets make this choice increasingly consequential. QM9 and ANI-1x span molecular chemical space, GAP-20 samples multiple forms of Carbon, revised MD17 resolves molecular trajectories, and OC20 covers heterogeneous catalytic environments~\cite{qm9,ani1x,gap20,rmd17,oc20}. Universal and foundation potentials extend this reach through pretraining, transfer and fine-tuning~\cite{m3gnet,chgnet,macemp,radova2025}. The resulting pools contain configurations with different degrees of redundancy and structural coverage, which can affect both apparent generalisation and transferability~\cite{redundancy,montesdeocazapiain2022selection}. For an already labelled pool, the selection problem is therefore how to use the available density-functional-theory (DFT) calculations effectively at the training-set size one intends to retain.

Coverage-based methods retain configurations that span the chosen representation of structure. Farthest-point sampling (FPS), determinantal point processes (DPP), QUESTS and minimum-set-cover variants select for geometric or information-theoretic diversity~\cite{fps,dpp,quests,questsmsc}. Clustering and stratified sampling pursue related goals, while batch active learning combines representativeness with scores used to acquire new configurations~\cite{qi2024,zaverkin2022}. Chemical and materials studies show that such space-filling strategies can be competitive and that the balance between exploration and exploitation changes with the available budget~\cite{gallagher2025dataefficiency,sivilotti2025highcoverage,schwalbekoda2025coverage}. Related work on submodular optimisation, coresets and gradient matching also seeks compact subsets that preserve information relevant to learning~\cite{submodular,craig,gradmatch,coresetsurvey}. Coverage provides a concrete starting point: retain a representative range of environments before asking a fitted model which regions need more attention.

A complementary strategy directs attention to configurations that expose model weakness. Active-learning workflows use uncertainty or extrapolation to guide new reference calculations during exploration and molecular dynamics~\cite{podryabinkin2017active,zhang2019dpgen,kulichenko2023uncertaintydriven,activelearning,flare,zaverkin2024,vitartas2026metadynamics}. These scores target regions where the current potential needs additional information. Their effectiveness depends on how uncertainty is estimated and combined with diversity; single-model estimates do not consistently outperform ensembles~\cite{ensembleuq,tan2023}. In an already labelled pool, model error and disagreement between models offer a related route to selecting configurations. Their value can then be tested directly by training on the selected subset and evaluating its predictions.

The difficulty is that a plausible selection score need not identify a more useful training subset. A large model error can mark an unfamiliar structure, so an error-based score may largely repeat the information already captured by coverage. A learned similarity measure can also preserve almost the same ordering as a fixed structural descriptor. Apparent missing-physics signals can likewise depend on the expressivity of that descriptor~\cite{pozdnyakov}. The unresolved issue is which differences between selection criteria translate into better force predictions, and at what retained-data budget. Answering it requires connecting the score used to choose configurations with the outcome of training on those configurations.

Retained-data budget can change the relative performance of selection rules. Data-pruning studies show that a criterion favoured under severe compression need not remain favoured when more data are retained~\cite{el2n,rho,pruning}; chemical selection studies report related budget-dependent trade-offs~\cite{gallagher2025dataefficiency,sivilotti2025highcoverage}. Structural coverage and model disagreement address different needs: spanning configuration space and focusing on discrepancies left by a trained model. Their balance can shift as the subset grows. A comparison at one retained fraction leaves that shift invisible. For atomistic compression, learning curves must therefore connect subset size to trained-model performance, while direct DFT errors establish whether reproducing a full-data model also improves prediction accuracy.

\begin{figure*}[!t]
\centering
\includegraphics[width=172mm]{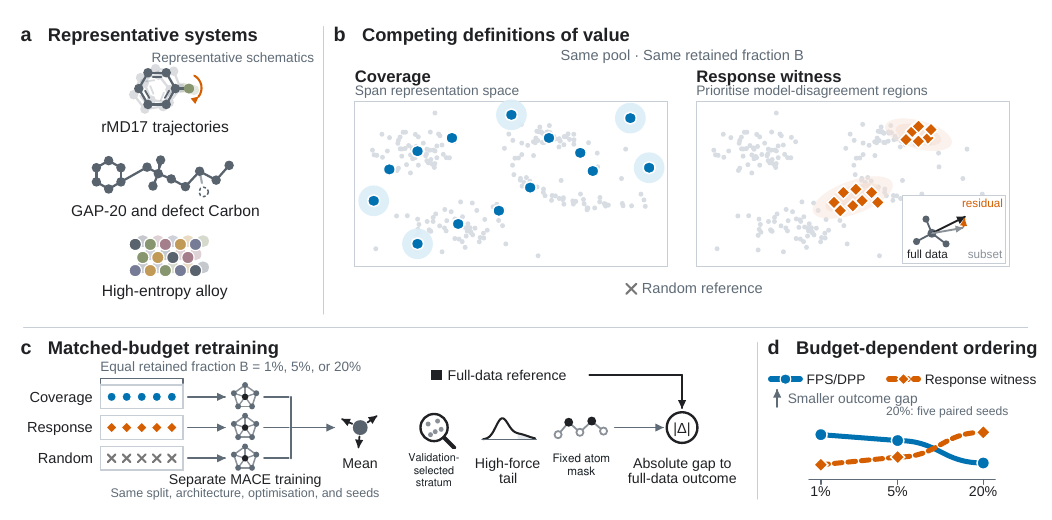}
\caption{\textbf{Matched-budget retraining separates structural coverage from response-guided data value.} \textbf{a}, Representative schematics for the rMD17 molecular trajectories, GAP-20 and defect-Carbon pools, and the high-entropy-alloy analysis. \textbf{b}, Competing selections from the same labelled pool at a fixed budget $B$: coverage favours diversity in representation space, the retrospective response witness favours disagreement with the full-data reference, and Random provides an unstructured control. \textbf{c}, Each selected subset trains a separate MACE model and is evaluated on held-out DFT force endpoints alongside the full-data reference. \textbf{d}, A schematic summary of the observed ordering introduces the central question: whether selector ordering changes between 1\%, 5\% and 20\% retained data. Quantitative estimates and paired-seed intervals are reported in Fig.~\ref{fig:budget_curves} and Tables~\ref{tab:paired_coverage} and~\ref{tab:response_crossover}.}
\label{fig:overview}
\end{figure*}

Here we address this question with a budget-resolved selection-and-retraining protocol on GAP-20 Carbon and pooled rMD17 (Fig.~\ref{fig:overview}). We compare Random, a validation-selected FPS/DPP coverage procedure and a response-guided selection at 1\%, 5\% and 20\% retained data. The response witness uses disagreement between a coverage-trained model and a full-data reference to prioritise regions for selection. Each selected subset then trains a separate MACE model under the same training protocol. Common data splits, force endpoints and paired training seeds make the selector comparisons directly interpretable. Learning curves measure agreement with the full-data outcome, and direct held-out DFT errors test the predictive benefit at the 20\% crossover.

We show that coverage and response selection exchange their ordering as the retained fraction increases. Coverage better reproduces the full-data outcome at 1\% and 5\%, whereas the response witness is closer at 20\% and also lowers direct DFT force error across all eight dataset--endpoint comparisons. The force endpoints reveal where response emphasis improves prediction and where it creates trade-offs. Complementary analyses of learned similarity, frozen-model error and the HEA25 energy target~\cite{hea,hea25data} connect these outcomes to the information represented by each selection score. Together, these contributions turn the choice between coverage and response into a testable decision at the intended training-set size, with predictive accuracy as the deciding evidence.

\section{Methods}
\label{sec:methods}

Our comparison links the rule used to select configurations to the force errors obtained after training. We vary selector and retained-data budget on common Carbon and rMD17 panels, using paired MACE training seeds and fixed force endpoints. Complementary descriptor analyses ask whether learned similarity changes the coverage ranking and whether a richer representation improves the HEA energy prediction. Foundation-error and response selections are tested through the predictions of models trained on their selected subsets.

\subsection{Atomistic datasets, splits and retained-data budgets}
\label{sec:methods_datasets}

The direct comparisons use fixed 1400-structure analysis panels derived from the complete generated training set in the GAP-20 Carbon archive~\cite{gap20,gap20data} and the ten revised MD17 (rMD17) molecular trajectories~\cite{rmd17,rmd17data}. For Carbon, 280 configurations were drawn from each of five prespecified physical groups: crystalline bulk and allotropes, liquid and amorphous Carbon, surfaces and interfaces, defects, and finite or low-dimensional structures. Sampling within every group covered within-group quintiles of energy per atom together with atom-count and volume-per-atom bins. Each group contributed 200 training, 40 validation and 40 held-out test structures. Its training allocation was further divided into 40 response-calibration and 160 selection-candidate structures.

For pooled rMD17, 140 frames were retained from each molecule. Every molecule contributed 100 training, 20 validation and 20 held-out test frames, with 20 training frames assigned to response calibration and 80 to the selection-candidate pool. Frames were ordered by the deposited \texttt{old\_indices}. Each trajectory was divided into non-overlapping temporal blocks using the larger decorrelation lag estimated from its energy-per-atom and RMS DFT-force series; at most one frame was sampled from a block, and a block could not contribute to more than one split.

For both datasets, configurations with non-finite energy or force labels were excluded, exact duplicate structures were assigned wholly to one split, and panel/split seed 0 was fixed before selection. The resulting 1000 training, 200 validation and 200 test structures were shared by every selector and training seed. Within training, the fixed 200-structure response-calibration partition was disjoint from the 800 selection candidates. Budgets of 10, 50 and 200 structures are named as 1\%, 5\% and 20\% of the 1000-structure full-data training reference; they correspond to 1.25\%, 6.25\% and 25\% of the eligible candidate pool. Results use paired training seeds 1--5. The source-data label ``rMD17 full'' denotes this complete pooled panel, not the full original archive. HEA25~\cite{hea,hea25data} supplies the non-spin energy-residual diagnostic, while MACE-MP-0~\cite{macemp} supplies frozen embeddings and force errors for the foundation-model analysis. Supplementary Table S0 gives the complete split contract and group definitions.

We study retrospective compression of already labelled pools. The response witness uses a full-data reference model to identify discrepancies that can guide the selection of a smaller training set.

\subsection{Coverage selectors and diagnostic scores}
\label{sec:methods_selectors}

Random selection without replacement is the unstructured reference. At each budget, validation chooses the primary coverage comparator between FPS and a DPP operating on the same standardised, mean-pooled invariant MACE-MP-0 descriptor embedding. The predeclared selection endpoint is normalised validation force MAE (lower is better), with tolerance $10^{-6}$ and FPS preferred before lexicographic structure ID on an exact tie. Carbon uses FPS at 1\% and DPP at 5\% and 20\%; rMD17 uses DPP at all three budgets. We use FPS/DPP as a compact family label in cross-budget figures. FPS starts from the structure farthest from the embedding centroid, and the DPP uses greedy maximum-a-posteriori selection from a radial-basis-function $L$-ensemble. Gaussian-kernel QUESTS coverage and its minimum-set-cover variant, QUESTS-MSC, provide additional coverage comparisons~\cite{quests,questsmsc,fps,dpp}. Algorithm settings and the retained coverage comparator at each budget are given in Supplementary Section~\ref{S-sec:si_selectors_criteria} and Table S0.

The DPP kernel bandwidth is the median positive pair distance, with diagonal jitter $10^{-10}$. The 800-structure candidate pools lie below the 4096-candidate threshold for the full kernel calculation. QUESTS-MSC uses bandwidth 0.015, 32 neighbours and a 5~\AA\ cutoff, held fixed across datasets and budgets. Each selection is constructed before outcome training; the five outcome seeds vary the model training, not the membership of the selected subset.

The four candidate signal families are: (i) a learned chemical/physical similarity; (ii) a descriptor-conditioned non-spin HEA energy residual; (iii) frozen MACE-MP-0 force error, used either directly or after residualisation; and (iv) a response witness derived from the discrepancy between a coverage model and the full-data reference. The learned-similarity and HEA studies examine the information carried by the descriptors. The foundation-error and response studies additionally test the value of the selected subsets through retraining. FPS/DPP is the primary coverage comparator, with QUESTS-MSC providing a secondary coverage comparison.

For learned similarity, each structure is assigned the physical attribute vector
\begin{equation}
\mathbf v_x=\left[\Delta e_x,\ \log(\overline{\lVert\mathbf F_x^{\mathrm{DFT}}\rVert}+\epsilon),\ \log\!\left(Q_{0.9}(\lVert\mathbf F_x^{\mathrm{DFT}}\rVert)+\epsilon\right)\right],
\label{eq:pair_attributes}
\end{equation}
where $e_x=E_x^{\mathrm{DFT}}/N_x$, $\Delta e_x$ is centred within the Carbon physical group, rMD17 molecule or HEA crystal--generation group, and the components are standardised on the metric-fit fold only. For fit-only pair $(i,j)$, the supervision target is
\begin{equation}
y_{ij}=\exp\!\left[-\frac{\lVert\widetilde{\mathbf v}_i-\widetilde{\mathbf v}_j\rVert_2^2}{2\ell^2}\right],
\label{eq:pair_target}
\end{equation}
where $\ell$ is the median positive pair distance in that fit fold. Let $c(x)$ denote coverage novelty and let $\mathbf{z}(x)$ collect force scale, density or composition and local structural descriptors. A candidate score $q(x)$ is represented as
\begin{equation}
q(x) \;=\; \mathbb{E}\!\left[q(x)\,\middle|\,c(x),\mathbf{z}(x)\right] \;+\; q^{\perp}(x),
\label{eq:cov_null}
\end{equation}
where $q^{\perp}$ is the part not explained by coverage and the fixed covariates. Rank correlation, top-budget Jaccard overlap, learned channel weights and leave-one-group-out AUROC describe whether the learned metric defines a distinct ranking. Pairs touching the held-out group are excluded from metric fitting, and the component scaling and pair-distance scale are refitted using the remaining groups. The pair-supervision objective and fixed-bank baseline are specified in Supplementary Section~\ref{S-sec:si_methods_detail}.

An auxiliary HEA25 analysis tests descriptor adequacy for non-spin DFT energy per atom, using grouped out-of-fold ridge regression after a fold-fitted elemental-composition baseline. The cumulative G0--G4 representations add radial, angular, ACE and frozen-MACE descriptors. Supplementary Eqs.~\ref{S-eq:si_hea_target} and~\ref{S-eq:si_hea_floor} define the target and normalised squared error; the complete descriptor, cross-validation and penalty-selection procedures are given there.

\subsection{MACE models and force-error endpoints}
\label{sec:methods_outcomes}

The Carbon and pooled rMD17 compression comparisons use the official MACE implementation~\cite{mace}, with the same architecture, loss, optimiser and epoch-based training schedule across selectors, retained-data budgets and full-data references. A separate model is fitted to each selected subset and compared with a reference trained on the 1000-structure training partition. Outcome seeds 1--5 are paired between subset and reference models. The frozen MACE-MP-0 representation used for coverage selection is distinct from these outcome models. Carbon-defect fine-tuning forms a separate foundation-model comparison.

The architecture used two interaction blocks, a 5.0~\AA\ cutoff, eight Bessel radial functions, a polynomial cutoff envelope with $p=5$, maximum spherical-harmonic degree $\ell_{\max}=3$ and correlation order 3. Hidden features were $128\times0e+128\times1o$. The weighted energy--force loss used weights 1 and 100. AdamW optimisation used an initial learning rate of 0.01, the standard MACE parameter-group weight-decay scheme with coefficient $5\times10^{-7}$, and batches of four structures. Training was configured for at most 1500 epochs with validation every epoch, ReduceLROnPlateau factor 0.8 and scheduler patience 50. Early-stopping patience was set to 200 epochs; exponential moving averaging used decay 0.99 and stage-two/SWA training was disabled. Supplementary Section~\ref{S-sec:si_mace_protocol} gives the remaining architecture and execution settings.

Training used CUDA and double precision, with energies in eV and forces in eV~$\text{\AA}^{-1}$. Atomic energy offsets, RMS-force scaling and the mean-neighbour normalisation were estimated separately from each model's training set, including the complete training partition for the full-data reference. Forces entered the training objective, whereas stress did not. Matched budget denotes equal retained structure count between selectors. The epoch-based schedule is shared across budgets, so the cross-budget curves do not hold the number of gradient updates fixed.

The 200-structure validation set is shared across selectors and budgets, and the 200-structure test set is reserved for held-out force evaluation. We assess subset performance both relative to the matched full-data model and directly against DFT labels. The HEA study separately measures out-of-fold energy prediction error across representations.

Let $\mathcal{D}$ be the 1000-structure full labelled training pool and let a selector return $S$ of size $B$ from the disjoint 800-structure candidate pool. Training on $S$ yields $\theta_S$, whereas training on all of $\mathcal{D}$ yields the full-data reference $\theta_{\mathcal{D}}$. Evaluation uses fixed, potentially overlapping groups $g$: the complete test set, the high-force tail, a validation-selected physical group and a high-novelty atom mask denoted in figures and tables as the High-novelty mask. Group definitions are frozen before test evaluation. Write $e_{xa}(\theta)=\lVert\mathbf F_{xa}^{\theta}-\mathbf F_{xa}^{\mathrm{DFT}}\rVert_2$ for the force-vector error of atom $a$ in structure $x$. The Mean, Validation-selected stratum and High-force tail endpoints use the arithmetic mean of $e_{xa}$ over their fixed atom sets, whereas the High-novelty mask endpoint uses the RMS of $e_{xa}$ over the high-novelty mask. Thus $L_g$ denotes the reported endpoint-specific force-error functional rather than a common MAE for every endpoint.
RMS aggregation gives larger atomwise errors more weight, making the masked endpoint sensitive to large local errors within the high-novelty region. We retain the sign of the relative deviation from the full-data reference and report its magnitude separately:
\begin{equation}
r_g(S)=100\,\frac{L_g(\theta_S)-L_g(\theta_{\mathcal D})}
{\max\{L_g(\theta_{\mathcal D}),\epsilon\}},
\qquad \delta_g(S)=|r_g(S)|.
\label{eq:signed_gap}
\end{equation}
Here $\epsilon=10^{-12}$ is a numerical guard. The learning curves show four absolute gaps $\delta_g$: Mean force error, Validation-selected stratum force error, High-force tail error and High-novelty mask RMS force error. The High-force tail is defined by the 80th percentile of the DFT force norm in the response-calibration partition (0.423621~eV~$\text{\AA}^{-1}$ for Carbon and 0.324778~eV~$\text{\AA}^{-1}$ for rMD17). The validation-selected group is frozen as
\begin{equation}
g^{\star}=\arg\max_{g}\frac{1}{5}\sum_{s=1}^{5}L_{g,\mathrm{val}}(\theta_{\mathcal D,s}),
\label{eq:validation_group}
\end{equation}
with lexicographic tie-breaking. Carbon uses the five physical groups defined above; its selected group is finite or low-dimensional Carbon. For the pooled rMD17 panel, the ten molecules are paired into five equal-size strata by increasing atom count, with ties ordered lexicographically; the selected second stratum contains benzene and uracil. The chosen group is transferred unchanged to every selector, budget and held-out test evaluation.

For the High-novelty mask, $\mathbf h_{xa}$ is the invariant atomic embedding from the frozen MACE-MP-0 medium model, and $n_{xa}=d_{5\mathrm{NN}}(\mathbf h_{xa},\mathcal H_{\mathcal K})$ is its five-nearest-neighbour distance to the response-calibration environment bank. Same-structure atoms are excluded when the calibration distribution is formed. The fixed mask is $M^{\mathrm{nov}}_{xa}=\mathbb{I}[n_{xa}\ge Q_{0.65}^{\mathcal K}(n)]$, with the calibration threshold projected unchanged to validation and test structures. This endpoint measures local force errors in structurally novel environments; trajectory drift is a separate dynamical quantity. The response witness is defined against $\theta_{\mathcal D}$, so agreement with its full-data outcome measures how well the selected subset reproduces the reference used in selection. Direct force errors $L_g$ against DFT labels establish the predictive benefit at the 20\% crossover.

\subsection{Foundation-error and response-based selections}
\label{sec:methods_response}

For the foundation-model analysis, the raw score is frozen MACE-MP-0 force error. Embedding novelty is the rank-normalised five-nearest-neighbour distance in the mean-pooled foundation embedding. Five-fold out-of-fold residualisation removes the variation explained by novelty and the fixed nuisance covariates. Raw-error and residual-error selections are then compared with embedding coverage and Random through one Carbon-defect fine-tuning endpoint: held-out force-vector MAE.

The response analysis uses a retrospective witness against the full-data model. For each budget $B$, let $S_{0,B}$ be the validation-selected FPS/DPP subset. A separate deterministic seed-0 pair of coverage and full-data models constructs the response selection, while seeds 1--5 are reserved for the paired outcome comparison. For structure $x$,
\begin{equation}
W_B(x) \;=\; \frac{1}{N_x}\sum_a \left\lVert \mathbf F_{\theta_{S_{0,B},0}}(x)_a-\mathbf F_{\theta_{\mathcal D,0}}(x)_a\right\rVert_2,
\label{eq:witness}
\end{equation}
Three disagreements distinguish energy shifts, distributed force mismatch and the largest local force mismatch. Write $\Delta E_{x,B}=E_{\theta_{S_{0,B},0}}(x)-E_{\theta_{\mathcal D,0}}(x)$ and $\Delta\mathbf F_{xa,B}=\mathbf F_{\theta_{S_{0,B},0}}(x)_a-\mathbf F_{\theta_{\mathcal D,0}}(x)_a$. The response coordinates are
\begin{equation}
\begin{aligned}
q_E(x)&=z_{\mathcal K}\!\left(|\Delta E_{x,B}|/N_x\right),\\
q_{F,\mathrm{RMS}}(x)&=z_{\mathcal K}\!\left(\sqrt{\frac{1}{N_x}\sum_a\lVert\Delta\mathbf F_{xa,B}\rVert_2^2}\right),\\
q_{F,\max}(x)&=z_{\mathcal K}\!\left(\max_a\lVert\Delta\mathbf F_{xa,B}\rVert_2\right).
\end{aligned}
\label{eq:response_coordinates}
\end{equation}
Each $z_{\mathcal K}(v)=(v-\mu_{\mathcal K})/\sigma_{\mathcal K}$ uses the coordinate's mean and population standard deviation from the fixed 200-structure response-calibration partition $\mathcal K$; a zero-variance coordinate is set to zero. Fixed edges $(-\infty,-1,0,1,+\infty)$ on each coordinate define 64 cells. These transformations and edges are projected unchanged to the disjoint 800 selection candidates.

For a cell with at least ten calibration structures, its priority is $R_B(c)=Q_{0.9}\{W_B(x):x\in\mathcal K,c(x)=c\}$. High-response cells have priorities at or above the 90th percentile of eligible-cell priorities. Occurrence is measured by candidate-pool atom mass, with the median among eligible cells as the threshold. A cell is occurred-but-not-learned (OBNL) if it meets both conditions and contains no member of $S_{0,B}$. The reported OBNL quantity is the fraction of candidate-pool atom mass assigned to these uncovered cells.

Global response-witness selection ranks candidates by their cell priority $R_B(c(x))$, not directly by the individual structure witness $W_B(x)$. Lexicographic structure ID breaks ties within a priority level. Response repair instead starts from the coverage subset and replaces 2\%, 5\% or 10\% of its members, rounded upward, with candidates from OBNL cells while preserving the total budget. The primary analysis uses the 5\% repair setting; when no eligible OBNL cell is present, the coverage subset is unchanged. Calibration fixes the response map, candidate structures supply the selectable configurations, and the held-out test set supplies the outcome errors. Supplementary Section~\ref{S-sec:si_methods_detail} gives the complete response-cell and endpoint definitions.

Response effects are reported separately for the Metadata-score mask, High-force tail, High-response cell and High-novelty mask errors. The Metadata-score mask is a composite physical-stress subset. For atom $a$ in structure $x$, its score is the mean of three calibration empirical ranks: the absolute deviation of $e_x$ from the calibration median in its physical group, the DFT force norm $\lVert\mathbf F^{\mathrm{DFT}}_{xa}\rVert_2$, and the embedding novelty $n_{xa}$ defined above. All medians, empirical-rank maps and the upper-quintile threshold are fitted on $\mathcal K$ and projected unchanged to the test set. For these 5\% comparisons, the reported effect is $100[L_g(\theta_{\mathrm{candidate}})-L_g(\theta_{\mathrm{FPS/DPP}})]/L_g(\theta_{\mathrm{FPS/DPP}})$, so positive values indicate higher error than FPS/DPP.

\subsection{Uncertainty estimation and paired comparisons}
\label{sec:methods_statistics}

Paired training-seed intervals are used for the primary selector contrasts. For each seed, $r_g$ and $\delta_g$ are calculated against the matched full-data model before averaging. All contrasts are calculated before rounding. At 5\%, we compare Random with FPS/DPP and response witness with FPS/DPP using absolute-gap differences. At 20\%, we compare witness with FPS/DPP using both absolute-gap and direct-error differences, and compare witness with the full-data reference using the signed gap. For each contrast, we enumerate all $5^5$ with-replacement resamples of the five paired differences and report the 2.5th and 97.5th percentiles. Paired $t$ intervals with four degrees of freedom provide a sensitivity check. These intervals measure training-seed variability conditional on the selected subsets.

Grouped analyses use 10,000 percentile-bootstrap resamples over the relevant scientific groups: five Carbon physical groups, ten rMD17 molecules or eight HEA25 FCC/BCC--generation-class groups. These intervals describe a different source of variation from the paired training-seed intervals. The 1\% comparison reports five-seed means. Supplementary Table~\ref{S-tab:si_interval_map} identifies the uncertainty estimate for each contrast.

\subsection{Manuscript and analysis-code preparation}

Generative-AI tools assisted language revision, methods exposition, numerical consistency checks and analysis and plotting-code preparation. Supplementary Section~\ref{S-sec:si_ai_assistance} describes these uses and representative preparation prompts; tool attribution is given in the Acknowledgments.

\begin{figure*}[!t]
\centering
\includegraphics[width=164mm]{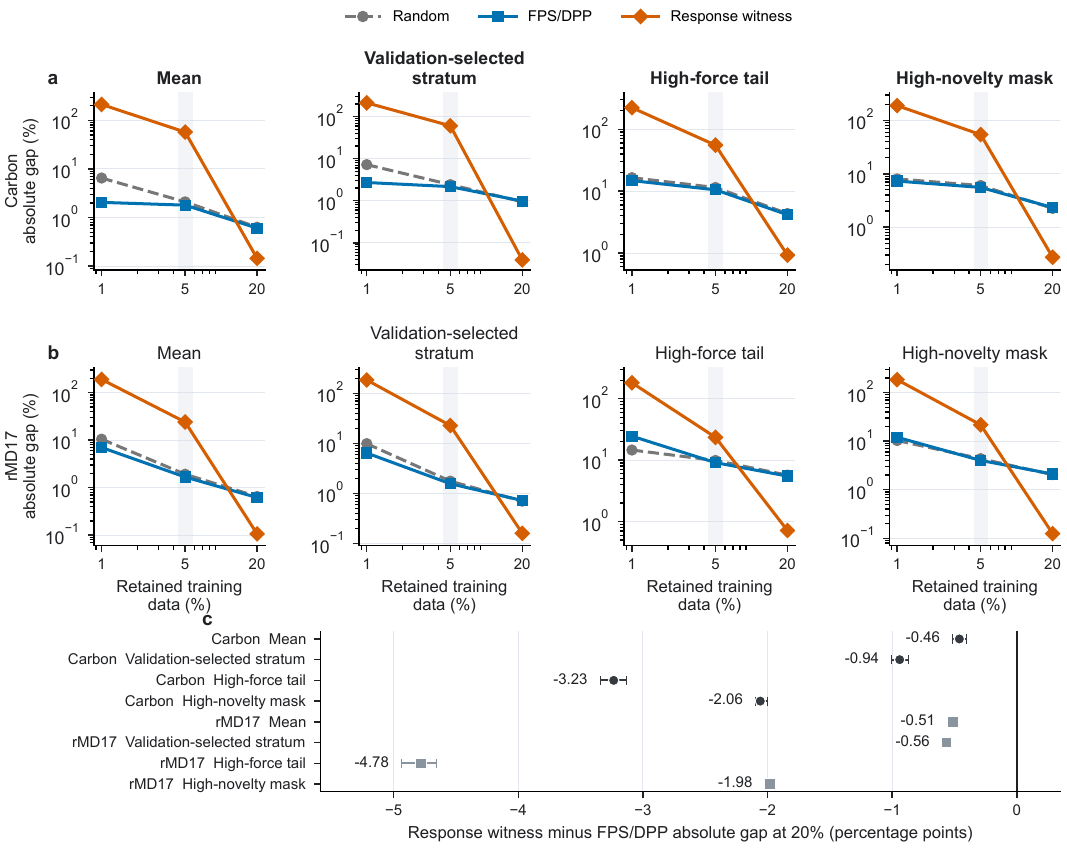}
\caption{\textbf{The response witness has larger absolute gaps at low budgets and smaller gaps at 20\%.} \textbf{a,b}, Absolute gaps to the full-data outcome for Carbon and pooled rMD17 at 1\%, 5\% and 20\%, for the Mean, Validation-selected stratum, High-force tail and High-novelty mask endpoints. Points are five-seed means and lines are visual guides. These panels show point estimates; paired uncertainty for the 20\% comparison is shown in \textbf{c}. \textbf{c}, Response-witness minus FPS/DPP absolute-gap differences at 20\%. Markers show five-seed means and error bars show 95\% percentile-bootstrap intervals from all paired resamples of the five seeds. Negative values indicate that the witness is closer to the full-data outcome. Table~\ref{tab:response_crossover} reports the direct held-out force errors.}
\label{fig:budget_curves}
\end{figure*}

\section{Results}
\label{sec:results}

Coverage and response selection show different dependence on retained-data budget. We first locate the reversal in their agreement with the full-data outcome, then test the 20\% response advantage against held-out DFT forces. The subsequent analyses examine which force regimes benefit and what information the selection scores carry.

\subsection{Selector ordering changes across retained-data budgets}
\label{sec:results_budget_ordering}

At the primary 5\% budget, FPS/DPP reproduced the full-data outcome more closely than Random across both datasets and all four endpoints (Table~\ref{tab:paired_coverage}). Its Mean absolute gap was 1.80\% in Carbon and 1.68\% in rMD17, compared with 2.10\% and 1.92\% for Random. The advantage also held for the Validation-selected stratum, High-force tail and High-novelty mask. Random minus FPS/DPP ranged from $+0.20$ to $+0.93$ percentage points (pp), with all eight paired training-seed intervals above zero. Supplementary Fig.~\ref{S-fig:si_outcome_gap} shows the individual seed values.

\begin{table*}[!t]
\centering
\small
\setlength{\tabcolsep}{6pt}
\caption{\textbf{FPS/DPP has smaller absolute gaps than Random at the 5\% budget.} Positive estimates mean that Random is farther from the full-data outcome. The 95\% intervals enumerate all paired percentile-bootstrap resamples of five matched training seeds and describe variation across those seeds.}
\label{tab:paired_coverage}
\begin{tabular}{llrr}
\toprule
Dataset & Endpoint & Random $-$ FPS/DPP (pp) & 95\% interval (pp) \\
\midrule
Carbon & Mean & $+0.303$ & $[+0.111,+0.411]$ \\
Carbon & Validation-selected stratum & $+0.251$ & $[+0.125,+0.329]$ \\
Carbon & High-force tail & $+0.931$ & $[+0.321,+1.291]$ \\
Carbon & High-novelty mask & $+0.470$ & $[+0.182,+0.640]$ \\
rMD17 full & Mean & $+0.244$ & $[+0.084,+0.356]$ \\
rMD17 full & Validation-selected stratum & $+0.197$ & $[+0.085,+0.263]$ \\
rMD17 full & High-force tail & $+0.769$ & $[+0.259,+1.079]$ \\
rMD17 full & High-novelty mask & $+0.350$ & $[+0.138,+0.478]$ \\
\bottomrule
\end{tabular}
\end{table*}

The absolute-gap ordering between coverage and the response witness reversed with budget (Fig.~\ref{fig:budget_curves}). At 1\% and 5\%, the witness had larger gaps for all four endpoints in both datasets. At 20\%, its Mean gap fell to 0.144\% in Carbon and 0.106\% in rMD17, compared with 0.605\% and 0.620\% for FPS/DPP. Witness minus FPS/DPP ranged from $-0.46$ to $-4.78$ percentage points across the eight dataset--endpoint pairs, with all paired-seed 95\% intervals below zero (Fig.~\ref{fig:budget_curves}c). The sampled budgets place this reversal between 5\% and 20\%.

The 20\% witness subsets also produced more accurate force predictions than FPS/DPP in all eight comparisons against held-out DFT labels (Table~\ref{tab:response_crossover}). Relative error reductions ranged from 0.46\% to 5.89\%, with all eight paired-seed intervals below zero. Mean force error decreased by 0.167~meV~$\text{\AA}^{-1}$ in Carbon and 0.164~meV~$\text{\AA}^{-1}$ in rMD17; the High-force tail and High-novelty mask showed larger reductions. The benefit at this budget therefore extends from reproducing the full-data outcome to improving predictive accuracy.

Six witness endpoints were also below the matched full-data reference: all four in rMD17 and the Carbon High-force tail and High-novelty mask. The Carbon Mean and Validation-selected stratum remained respectively 0.144\% and 0.039\% above full-data error. Supplementary Table~\ref{S-tab:si_20pct_paired} reports the paired-seed intervals for these signed gaps. Thus models trained on the witness subsets exceeded the full-data reference on six force endpoints under the shared training protocol. Across budgets, the absolute-gap curves establish the change in reference agreement; at 20\%, the direct DFT errors establish the accuracy gain over coverage.

\begin{table*}[t]
\centering
\small
\setlength{\tabcolsep}{3.5pt}
\caption{\textbf{Response-witness selection lowers direct held-out force error at the 20\% budget.} Errors are five-seed means in meV~$\text{\AA}^{-1}$. The signed gap $r_g$ is the percentage deviation of the witness error from the matched full-data error. Relative gaps are calculated within each seed before averaging and need not equal ratios of the displayed mean errors. Differences are response witness minus FPS/DPP. Brackets give 95\% percentile-bootstrap intervals from all paired resamples of five matched seeds. Negative differences favour the witness over FPS/DPP; negative $r_g$ indicates lower error than full data. Supplementary Table~\ref{S-tab:si_20pct_paired} gives the seed values and paired-$t$ sensitivity intervals.}
\label{tab:response_crossover}
\begin{tabular}{llrrrrr}
\toprule
Dataset & Endpoint & Full data & FPS/DPP & Witness & Witness $r_g$ (95\% CI), \% & Difference (95\% CI) \\
\midrule
Carbon & Mean & 36.162 & 36.380 & 36.214 & $+0.144$ $[+0.127,+0.161]$ & $-0.167$ $[-0.186,-0.146]$ \\
Carbon & Validation-selected stratum & 34.605 & 34.944 & 34.619 & $+0.039$ $[+0.019,+0.059]$ & $-0.325$ $[-0.347,-0.301]$ \\
Carbon & High-force tail & 37.015 & 38.555 & 36.672 & $-0.927$ $[-1.132,-0.763]$ & $-1.883$ $[-2.050,-1.745]$ \\
Carbon & High-novelty mask & 40.535 & 41.480 & 40.425 & $-0.273$ $[-0.308,-0.240]$ & $-1.056$ $[-1.099,-1.007]$ \\
rMD17 full & Mean & 22.519 & 22.659 & 22.495 & $-0.106$ $[-0.139,-0.073]$ & $-0.164$ $[-0.183,-0.146]$ \\
rMD17 full & Validation-selected stratum & 22.817 & 22.982 & 22.781 & $-0.160$ $[-0.188,-0.133]$ & $-0.201$ $[-0.217,-0.186]$ \\
rMD17 full & High-force tail & 23.958 & 25.275 & 23.787 & $-0.716$ $[-0.760,-0.669]$ & $-1.488$ $[-1.524,-1.456]$ \\
rMD17 full & High-novelty mask & 24.219 & 24.729 & 24.189 & $-0.125$ $[-0.153,-0.102]$ & $-0.540$ $[-0.557,-0.526]$ \\
\bottomrule
\end{tabular}
\end{table*}

Random and coverage showed mixed absolute-gap ordering under the strongest compression. At 1\% in rMD17, FPS/DPP had smaller Mean and Validation-selected stratum gaps, whereas Random had smaller High-force tail and High-novelty mask gaps. With ten structures retained, the subset can leave some molecules unrepresented. QUESTS-MSC also had large low-budget Mean gaps, reaching 42.98\% for Carbon and 95.16\% for rMD17. Its fixed parameterisation used bandwidth 0.015, 32 neighbours and a 5~\AA\ cutoff without dataset- or budget-specific tuning. These results further illustrate how severe compression exposes differences between selection rules and force endpoints.

\subsection{Response emphasis creates endpoint-specific trade-offs at 5\%}
\label{sec:results_response_repair}

\begin{figure*}[!t]
\centering
\includegraphics[width=154mm]{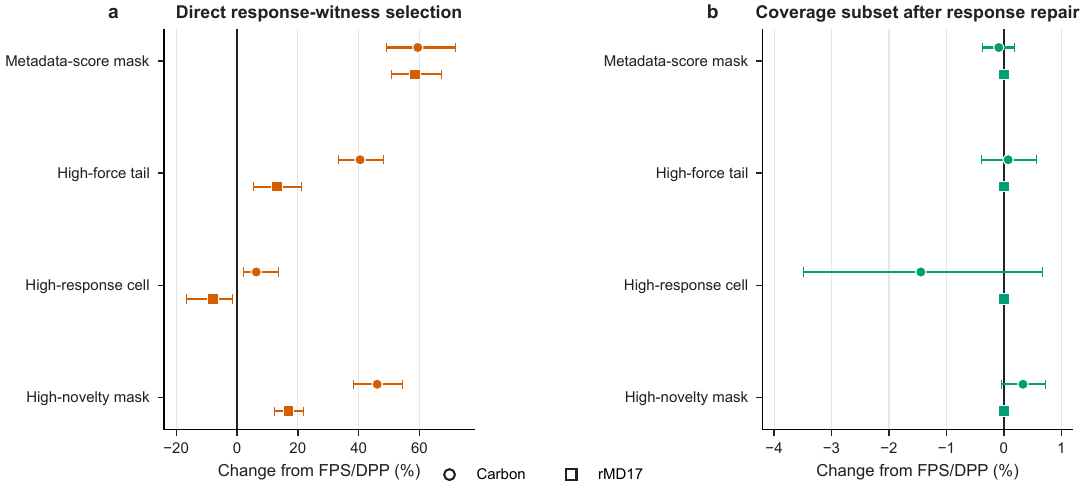}
\caption{\textbf{Response emphasis improves one local rMD17 endpoint but increases the other reported errors at 5\%.} \textbf{a}, Response-witness effects relative to FPS/DPP across four held-out force-error endpoints. \textbf{b}, Response-repair effects relative to FPS/DPP. Negative values indicate lower error; error bars are grouped-bootstrap 95\% confidence intervals. Endpoint masks can overlap and are not independent replications. The horizontal scales differ between panels. Carbon repair replaces three of 50 structures, reduces candidate-pool OBNL atom mass from 1.00\% to zero and retains a Jaccard overlap of 0.887. rMD17 repair returns the coverage subset because FPS/DPP already covers every eligible high-response cell.}
\label{fig:response_repair}
\end{figure*}

At 5\%, response-witness selection improved one local rMD17 endpoint while worsening the other reported errors (Fig.~\ref{fig:response_repair}a). The High-response cell error fell by 7.87\%, but the Metadata-score mask, High-force tail and High-novelty mask errors increased by 13.20--58.52\%. In Carbon, all four errors increased, from $+6.33\%$ for the High-response cell to $+59.45\%$ for the Metadata-score mask; all four grouped-bootstrap intervals excluded zero. Response emphasis therefore produced a local gain in rMD17 without a uniform improvement across its evaluation regimes.

Response repair made a smaller change to the coverage subset (Fig.~\ref{fig:response_repair}b). In Carbon, replacing 3 of 50 structures, a realised replacement fraction of 6\%, reduced candidate-pool OBNL atom mass from 1.00\% to zero. The repaired subset retained a Jaccard overlap of 0.887 with FPS/DPP. Coverage already included every well-populated high-response cell in rMD17, so repair made no replacement and returned the same subset.

Closing the Carbon coverage gap did not produce a detectable error reduction. Repair changed the Metadata-score mask, High-force tail, High-response cell and High-novelty mask errors by $-0.088\%$, $+0.076\%$, $-1.446\%$ and $+0.333\%$, respectively, with every 95\% interval including zero. The rMD17 effects were exactly zero because its subset was unchanged. Supplementary Table~\ref{S-tab:si_response_repair_summary} gives the estimates and intervals. The intervention changed response-cell coverage without a resolved improvement in these force endpoints.

The five-seed means at 1\% show a different repair trade-off (Supplementary Table~\ref{S-tab:si_budget_sensitivity}). In Carbon, repair reduced the High-force tail absolute gap from 14.80\% to 7.66\% and the high-novelty masked gap from 7.30\% to 3.90\%, while increasing the Mean gap from 2.05\% to 2.57\% and the Validation-selected stratum gap from 2.72\% to 2.92\%. In rMD17, repair reduced all four mean gaps by 0.77--2.66~pp. These absolute-gap changes measure how closely the subset models reproduce the full-data outcome. The contrast with 5\% shows that even a small repair of a coverage subset has a budget- and endpoint-dependent effect.

\subsection{Frozen foundation-model error largely tracks embedding novelty}
\label{sec:results_foundation_error}

\begin{figure*}[!t]
\centering
\includegraphics[width=160mm]{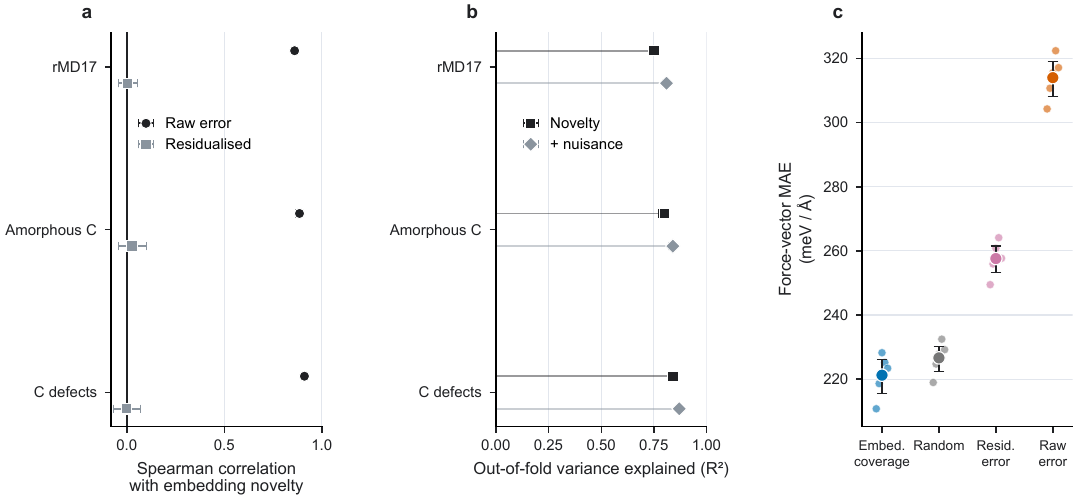}
\caption{\textbf{Frozen-model error tracks embedding novelty, and error-based selections give higher Carbon-defect MAE than coverage.} \textbf{a}, Spearman correlations of raw and residualised force error with five-nearest-neighbour novelty in the mean-pooled MACE-MP-0 embedding. \textbf{b}, Out-of-fold variance explained by novelty alone and by novelty plus nuisance covariates. Points and error bars in \textbf{a} and \textbf{b} show estimates and grouped-bootstrap 95\% confidence intervals. \textbf{c}, Held-out Carbon-defect force-vector MAE for five training seeds. Small points show individual seeds; large symbols and error bars show five-seed means and reported 95\% confidence intervals.}
\label{fig:foundation_error}
\end{figure*}

Frozen MACE-MP-0 force error closely tracked novelty in the same model's embedding, limiting its distinction from a coverage score (Fig.~\ref{fig:foundation_error}a,b). Spearman correlations were 0.86 for rMD17, 0.89 for amorphous Carbon and 0.91 for Carbon defects. Novelty alone explained 0.75--0.84 of the out-of-fold error variance; adding the fixed nuisance covariates increased this range to 0.81--0.87.

Five-fold out-of-fold residualisation reduced these correlations to 0.004 for rMD17, 0.026 for amorphous Carbon and $-0.002$ for Carbon defects. The residual score therefore removed most of the rank association with novelty, allowing a downstream comparison of error information beyond that association.

Embedding coverage remained the best of these selections for Carbon-defect force-vector MAE (Fig.~\ref{fig:foundation_error}c). Its five-seed mean held-out error was 221.2~meV~$\text{\AA}^{-1}$, compared with 226.6 for Random, 257.6 for residual-error selection and 314.1~meV~$\text{\AA}^{-1}$ for raw-error selection. The corresponding paired differences were $+5.4$, $+36.4$ and $+92.8$~meV~$\text{\AA}^{-1}$, with all three reported intervals above zero. Removing novelty from frozen-model error therefore did not make direct error-based selection more effective for this endpoint.

\subsection{Learned similarity preserves the coverage ranking}
\label{sec:results_learned_similarity}

\begin{figure*}[!t]
\centering
\includegraphics[width=160mm]{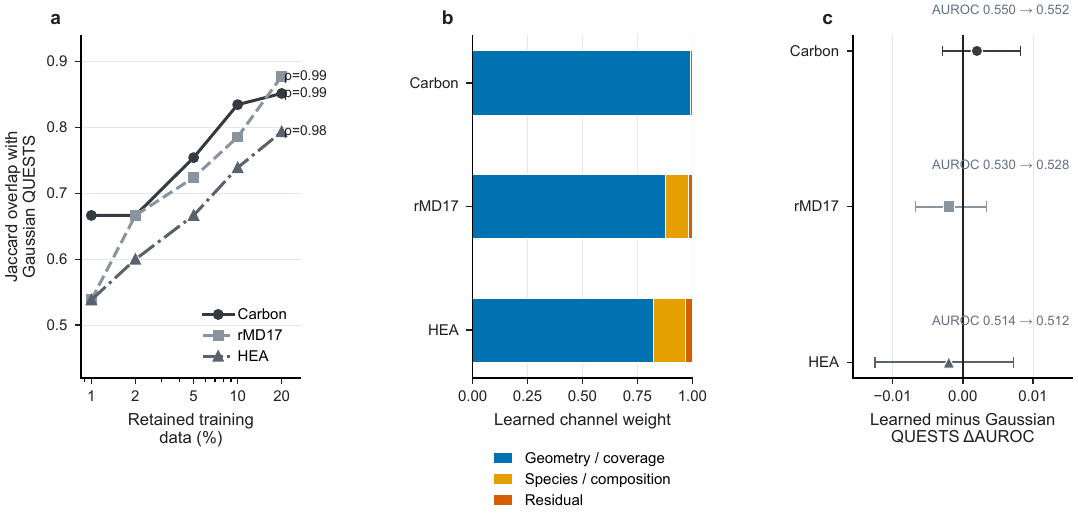}
\caption{\textbf{Learned chemical and physical similarity preserves the coverage ranking.} \textbf{a}, Selected-set overlap with Gaussian QUESTS across retained-data budgets; annotations report full-ranking Spearman correlations. \textbf{b}, Learned contributions from geometry and coverage, species and composition, and the residual channel. \textbf{c}, Learned-minus-Gaussian leave-one-group-out $\Delta$AUROC with grouped-bootstrap 95\% confidence intervals. All three intervals include zero.}
\label{fig:learned_similarity}
\end{figure*}

Learning chemical and physical similarity also retained a ranking close to coverage across the three diagnostic datasets (Fig.~\ref{fig:learned_similarity}). Spearman correlations with Gaussian QUESTS ranged from 0.980 to 0.990, and selected-set Jaccard overlap at 5\% ranged from 0.667 to 0.754 (Fig.~\ref{fig:learned_similarity}a). Geometry and coverage accounted for 0.821--0.989 of the fitted weight, whereas the residual channel received no more than 0.030 (Fig.~\ref{fig:learned_similarity}b). The fitted weights and ranking agreement thus pointed to the same dominant source of information.

The ranking agreement was accompanied by similar group-discrimination performance. Absolute leave-one-group-out AUROCs were close to chance, and learned-minus-Gaussian $\Delta$AUROC intervals included zero for Carbon, rMD17 and HEA (Fig.~\ref{fig:learned_similarity}c). In the constructed positive control, the same parameterisation departed from coverage when an independent channel was injected. The model could therefore respond to that controlled signal, while the measured physical attributes produced rankings dominated by structural coverage.

\section{Discussion}
\label{sec:discussion}

\subsection{Data budget is part of selector performance}

Retained-data budget changes the balance between structural coverage and model response. Coverage gives smaller absolute gaps to the full-data outcome than the response witness at 1\% and 5\%, whereas the witness gives smaller gaps in all eight dataset--endpoint comparisons at 20\%. At 20\%, the witness advantage is also present in direct held-out DFT force errors across all eight comparisons. Choosing a selection rule therefore requires specifying how much data will be retained and which prediction errors matter.

This crossover connects atomistic training-set selection with data pruning. The preferred pruning criterion can change as the retained fraction increases~\cite{pruning}, and chemical selection studies report related budget-dependent exploration--exploitation trade-offs~\cite{gallagher2025dataefficiency,sivilotti2025highcoverage}. Our force-based MLIP comparisons locate the change between the sampled 5\% and 20\% budgets. Intermediate learning-curve points would resolve whether the transition is gradual or sharp.

Model-dependent example value helps explain why the ranking can change. Influence, forgetting, training dynamics and early-loss rankings measure different aspects of how a model uses an example~\cite{influence,forgetting,cartography,el2n,rho}. Structural coverage and response disagreement emphasise different requirements: spanning configuration space and targeting discrepancies left by a trained model. As the subset grows, the balance between these requirements can shift. The observed crossover makes this balance a concrete question for atomistic training-set design.

\subsection{Response emphasis exposes endpoint trade-offs}

The 5\% response experiments show that a gain in one force regime can accompany losses elsewhere. In rMD17, the witness lowers High-response cell error while increasing Metadata-score mask, High-force tail and High-novelty mask errors. In Carbon, every reported response endpoint worsens. This endpoint dependence is relevant to materials benchmarks, where the target property and evaluation protocol affect model comparisons~\cite{matbenchdiscovery}. A selection that targets a local discrepancy must therefore also be evaluated on the broader force distribution.

Response repair tests the effect of a small change in subset membership. In Carbon it removes the 1.00\% uncovered high-response atom mass, yet all outcome intervals include zero. In rMD17, the coverage subset already contains every eligible high-response cell and the reported repair subset is unchanged. These results separate coverage of a response-defined region from measurable improvement after retraining.

At 20\%, the witness improves direct DFT force errors as well as agreement with the full-data outcome. All eight paired-seed comparisons favour the witness over FPS/DPP, and six errors are also below the full-data reference. Their signed-gap intervals remain below zero (Table~\ref{tab:response_crossover}), as do the paired-$t$ sensitivity intervals in Supplementary Table~\ref{S-tab:si_20pct_paired}. The Carbon Mean and Validation-selected stratum endpoints are 0.144\% and 0.039\% higher, respectively. A response-selected subset can therefore yield more accurate predictions than training on all 1000 structures for particular force regimes under the shared training protocol. A matched-update comparison would test whether this ordering persists at equal numbers of optimisation steps.

The immediate application is compression of an already labelled pool using information from a coverage-trained model and a full-data reference. The demonstrated benefit is lower held-out force error at the 20\% retained-data budget. End-to-end computational savings would additionally depend on the cost of constructing these preliminary models and training the selected subsets. Prospective acquisition poses a different design task: estimating useful response information before new DFT labels are computed, as in workflows that combine uncertainty, diversity and iterative exploration~\cite{podryabinkin2017active,zhang2019dpgen,kulichenko2023uncertaintydriven,zaverkin2022,zaverkin2024,vitartas2026metadynamics}.

\subsection{Selection scores and training value are different quantities}

The complementary analyses explain why selection scores need to be connected to trained-model outcomes. Learned similarity retains Spearman correlations of 0.98--0.99 with Gaussian QUESTS and assigns most weight to geometry and coverage. Frozen-model error follows embedding novelty, while direct error-based selection increases Carbon-defect force MAE relative to embedding coverage. These are two different outcomes: a score can reproduce an existing ranking, or it can change the selection without improving prediction. Retraining distinguishes useful information from either case.

The auxiliary HEA25 regression illustrates a related dependence on representation. The G4 descriptors reduce normalised out-of-fold squared error to 0.240 of the G0 value, with a grouped-bootstrap interval of $[0.222,0.278]$ (Supplementary Table~\ref{S-tab:si_residual_gap}). This regression tests how well the descriptors represent the energy target. Rank and overlap statistics test whether selection scores order structures differently, and matched-budget retraining tests whether those differences improve prediction. The same sequence can be used to evaluate ensemble uncertainty and hybrid diversity--uncertainty strategies~\cite{ensembleuq,tan2023}.

\subsection{Implications for atomistic data selection}

A practical comparison starts with coverage and Random, evaluates candidate selectors at several matched budgets, and reports both direct force errors and deviations from a full-data reference. Multiple endpoints reveal whether improvements are concentrated in the mean, tail or a particular local regime. Uncertainty should reflect the quantity being compared: paired training-seed intervals measure optimisation variability, whereas grouped resampling measures variation across the specified scientific groups.

Our comparison uses a validation-selected FPS/DPP procedure and a single Random subset at each budget, with intervals describing training-seed variability for those selected subsets. The three-budget curves span five training seeds, two atomistic panels and one MACE-MP-0 coverage representation. They show why choosing a selector and choosing a training-set size are coupled decisions. A method favoured under strong compression can give way to another at a larger retained fraction, so the relevant comparison is the one made at the intended budget and force endpoint.

\section{Conclusions}
\label{sec:conclusions}

The amount of data retained changes the relative performance of coverage- and response-based training-set selection. Our selection-and-retraining comparison on GAP-20 Carbon and pooled rMD17 reveals a crossover: FPS/DPP is closer to the full-data outcome than Random at 5\% across all eight force endpoints, and the response witness has larger absolute gaps than coverage at 1\% and 5\%, but outperforms coverage at 20\%. At 20\%, witness-selected models also reduce direct held-out force errors by 0.46--5.89\% relative to FPS/DPP, with all eight paired-seed intervals favouring the witness; six errors are below the full-data reference.

The complementary analyses identify why an appealing score can fall short as a selection criterion. Learned similarity preserves the coverage ranking, and frozen-model-error selection increases Carbon-defect force MAE relative to embedding coverage. Connecting selection scores to retrained-model predictions across budgets therefore provides the decisive test of data value. The resulting design lesson is direct: select the training-set size and selection rule together, and judge the chosen subset by the force accuracy required of the final model.

\section*{Author contributions}
J.B.: Conceptualization, Methodology, Software, Investigation, Formal analysis, Data curation, Visualization, Writing -- original draft. A.M.E.: Data curation, Software, Validation, Writing -- review and editing. Both authors discussed the results and approved the manuscript.

\section*{Conflicts of interest}
There are no conflicts to declare.

\section*{Data availability}
The GAP-20 Carbon~\cite{gap20,gap20data}, revised MD17 (rMD17)~\cite{rmd17,rmd17data} and HEA25~\cite{hea,hea25data} datasets are publicly available. The pretrained model used for embeddings and error estimates is MACE-MP-0~\cite{macemp}. The accompanying source-data and analysis archive is associated with the Zenodo record \url{https://doi.org/10.5281/zenodo.22263567}~\cite{outcomeauditdata}. It includes five-seed aggregate values for the 1\%, 5\% and 20\% budget curves, seed-resolved values for the primary 5\% comparisons and 20\% response crossover, and the manuscript source files. A portable analysis entry point regenerates the paired statistics and quantitative figures from the CSV tables without training models. The archive also contains a MACE configuration specifying the shared compression-training settings and a data dictionary. The companion development repository is \url{https://github.com/JIABI/QUESTS_physical_kernel}. Supplementary Section~\ref{S-sec:si_reproducibility_manifest} maps the archived source data and code.

\section*{Acknowledgments}
The authors were supported by the Ada Lovelace Centre at the Science and Technology Facilities Council (
https://adalovelacecentre.ac.uk). A.M.E was also supported by the Physical Sciences Data Infrastructure (https://psdi.ac.uk; jointly STFC and the University of Southampton) under grants EP/X032663/1 and EP/X032701/1, and EPSRC under grants EP/W026775/1 and EP/V028537/1.
 We acknowledge computational resources from STFC Scientific Computing Department's SCARF cluster and cloud.





\bibliography{rsc} 
\bibliographystyle{rsc} 
\end{document}


\raggedbottom

\pagestyle{fancy}
\thispagestyle{plain}
\fancypagestyle{plain}{
\renewcommand{\headrulewidth}{0pt}
}

\makeFNbottom
\makeatletter
\renewcommand\LARGE{\@setfontsize\LARGE{15pt}{17}}
\renewcommand\Large{\@setfontsize\Large{12pt}{14}}
\renewcommand\large{\@setfontsize\large{10pt}{12}}
\renewcommand\footnotesize{\@setfontsize\footnotesize{7pt}{10}}
\makeatother

\renewcommand{\thefootnote}{\fnsymbol{footnote}}
\renewcommand\footnoterule{\vspace*{1pt}%
\color{cream}\hrule width 3.5in height 0.4pt \color{black}\vspace*{5pt}} 
\setcounter{secnumdepth}{5}

\makeatletter 
\renewcommand\@biblabel[1]{#1}            
\renewcommand\@makefntext[1]%
{\noindent\makebox[0pt][r]{\@thefnmark\,}#1}
\makeatother 
\renewcommand{\figurename}{\small{Fig.}~}
\renewcommand{\thefigure}{S\arabic{figure}}
\renewcommand{\thetable}{S\arabic{table}}
\renewcommand{\theequation}{S\arabic{equation}}
\sectionfont{\sffamily\Large}
\subsectionfont{\normalsize}
\subsubsectionfont{\bf}
\setstretch{1.125} 
\setlength{\skip\footins}{0.8cm}
\setlength{\footnotesep}{0.25cm}
\setlength{\jot}{10pt}
\titlespacing*{\section}{0pt}{4pt}{4pt}
\titlespacing*{\subsection}{0pt}{15pt}{1pt}

\fancyfoot{}
\fancyfoot[CO]{\footnotesize\sffamily Digital Discovery, [year], [vol.],}
\fancyfoot[CE]{\footnotesize\sffamily Digital Discovery, [year], [vol.],}
\fancyfoot[RO]{\footnotesize{\sffamily{1--\pageref{LastPage} ~\textbar  \hspace{2pt}\thepage}}}
\fancyfoot[LE]{\footnotesize{\sffamily{\thepage~\textbar\hspace{3.45cm} 1--\pageref{LastPage}}}}
\fancyhead{}
\renewcommand{\headrulewidth}{0pt} 
\renewcommand{\footrulewidth}{0pt}
\setlength{\arrayrulewidth}{1pt}
\setlength{\columnsep}{6.5mm}
\setlength\bibsep{1pt}

\makeatletter 
\newlength{\figrulesep} 
\setlength{\figrulesep}{0.5\textfloatsep} 

\newcommand{\topfigrule}{\vskip-1pt{\color{cream}\hrule height 1.5pt}\vskip\figrulesep}

\newcommand{\botfigrule}{\vskip\figrulesep{\color{cream}\hrule height 1.5pt}\vskip-2pt}

\newcommand{\dblfigrule}{\vskip-1pt{\color{cream}\hrule height 1.5pt}\vskip\figrulesep}

\makeatother

\twocolumn[
  \begin{@twocolumnfalse}
 {\raisebox{8pt}{\sffamily\bfseries\fontsize{18}{20}\selectfont Digital Discovery}\hfill\raisebox{0pt}[0pt][0pt]{\includegraphics[height=55pt]{head_foot/RSC_LOGO_CMYK}}\\[1ex]
{\setlength{\fboxsep}{0pt}\colorbox{ddblue}{\parbox[c][18pt][c]{18.5cm}{\hspace{0.45cm}\color{white}\sffamily\bfseries\fontsize{8}{9}\selectfont SUPPLEMENTARY INFORMATION}}}}\par
\vspace{1em}
\sffamily
\begin{tabular}{@{}m{4.5cm} p{13.5cm}@{}}

\includegraphics{head_foot/DOI} & \noindent\LARGE{\textbf{Supplementary Information for: A budget-dependent crossover between coverage- and response-based training-set selection for machine-learned interatomic potentials}} 
\\
\vspace{0.3cm} & \vspace{0.3cm} \\

 & \noindent\large{Jia Bi,$^{\ast}$\textit{$^{a}$} Alin-Marin Elena,$^{\ast}$\textit{$^{b}$}} \\

\includegraphics{head_foot/dates} & \noindent\normalsize{This document provides extended methods, diagnostic controls, five-seed outcome summaries, budget-dependent learning curves and a constructed methodological positive control. The corresponding source-data tables are archived at \url{https://doi.org/10.5281/zenodo.22263567}.} \\

\end{tabular}

 \end{@twocolumnfalse} \vspace{0.6cm}

  ]

\renewcommand*\rmdefault{bch}\normalfont\upshape
\rmfamily
\section*{}
\vspace{-1cm}


\footnotetext{\textit{$^{a}$~Science and Technology Facilities Council, Harwell Campus, Didcot, United Kingdom. E-mail: Jia.Bi@stfc.ac.uk}}
\footnotetext{$^{\ast}$~Corresponding author. \textit{$^{b}$~Science and Technology Facilities Council, Keckwick Lane, Daresbury, United Kingdom. E-mail: alin-marin.elena@stfc.ac.uk}}



\section{Methods}

The numerical results in this Supplementary Information are provided as source tables at \url{https://doi.org/10.5281/zenodo.22263567}. Main-text tables are organised under \texttt{data/main/}, and Supplementary Information tables under \texttt{data/si/}. The same tables drive the submitted figures and the numerical cross-checks described below.

\subsection{Datasets, splits and budgets}
\label{sec:si_datasets_splits}

The direct outcome analyses use fixed 1400-structure panels from GAP-20 Carbon~\cite{gap20,gap20data} and the ten rMD17 trajectories~\cite{rmd17,rmd17data}. All data roles were assigned before selector fitting. Non-finite energy or force labels were excluded. An exact-geometry duplicate key comprised the ordered atomic numbers, periodic boundary conditions, cell and coordinates at the precision deposited by the source archive; energy and force labels were not part of the key. All members of a duplicate group were assigned to the same split. Panel/split seed 0 was used once, and the resulting panel and split were shared by every selector and training seed.

The GAP-20 panel was sampled from the complete generated Carbon set, \texttt{Carbon\_Data\_Set\_Total.xyz} in the unpacked archive. The native \texttt{config\_type} labels were mapped to five physical groups listed in Table S0. Within each group, energy per atom, atom count and volume per atom ($|\det\mathbf H_x|/N_x$ from the deposited cell) were divided into empirical quintile bins; tied quantile edges were merged. The fixed-seed stratified draw covered the non-empty combinations of these bins and retained 280 configurations per physical group. Each group contributed 200 training, 40 validation and 40 held-out test configurations. Within its training allocation, 40 structures formed the response-calibration partition and 160 formed the candidate pool.

The rMD17 panel retained 140 frames from each of aspirin, azobenzene, benzene, ethanol, malonaldehyde, naphthalene, paracetamol, salicylic acid, toluene and uracil. For molecule $m$, frames were ordered by \texttt{old\_indices}. The block width was $\tau_m=\max(\tau_{E,m},\tau_{F,m})$, where $\tau_{E,m}$ and $\tau_{F,m}$ are the first positive lags at which the sample autocorrelation of energy per atom and per-structure RMS DFT-force norm, respectively, is no greater than $e^{-1}$. Non-overlapping blocks were formed in this order, no more than one frame was drawn from a block, and one block could not cross data roles. Every molecule contributed 100 training, 20 validation and 20 test frames; 20 of its training frames formed the response-calibration partition and 80 formed the candidate pool. The stable record for a frame is the tuple (molecule, \texttt{old\_indices}, block ID, split, response role).

Both panels therefore contain 1000 training, 200 validation and 200 held-out test structures. The full-data reference uses all 1000 training structures, including the 200 calibration structures. Subset selectors can choose only from the disjoint 800-structure candidate pool. The reported subset sizes of 10, 50 and 200 are 1\%, 5\% and 20\% of the full-data training reference and 1.25\%, 6.25\% and 25\% of the eligible candidate pool. The label ``rMD17 full'' denotes this complete pooled analysis panel, not the full original rMD17 archive. HEA25~\cite{hea,hea25data} is used for learned-similarity and non-spin energy-residual diagnostics; Carbon defects provide the foundation-model fine-tuning endpoint.

\begin{table*}[!tp]
\centering
\small
\caption*{\textbf{Supplementary Table S0. Dataset construction, split and validation contract.} Response calibration is contained within the training pool, whereas subset selection is restricted to the candidate pool. The validation-selected stratum maximises the five-seed full-reference validation force error and is then frozen for every held-out comparison. Selection of the coverage comparator minimises normalised validation force MAE; exact ties favour FPS and then lexicographic structure ID.}
\noindent\textit{Panel A: global split and budget contract.}\\[2pt]
\begin{tabular}{lrrrrcc}
\toprule
Dataset & Train & Calibration & Candidates & Validation & Test & Budgets \\
\midrule
Carbon & 1000 & 200 & 800 & 200 & 200 & 10/50/200 \\
rMD17 full & 1000 & 200 & 800 & 200 & 200 & 10/50/200 \\
\bottomrule
\end{tabular}

\vspace{0.7em}
\noindent\textit{Panel B: Carbon physical groups and native GAP-20 labels.}\\[2pt]
\adjustbox{max width=\textwidth}{%
\begin{tabular}{lll}
\toprule
Code & Physical group & Native \texttt{config\_type} labels \\
\midrule
STRATUM\_1 & Crystalline bulk/allotropes & Crystalline\_Bulk, Crystalline\_RSS, Diamond, Graphite, SACADA \\
STRATUM\_2 & Liquid/amorphous & Amorphous\_Bulk, Liquid \\
STRATUM\_3 & Surfaces/interfaces & Amorphous\_Surfaces, Liquid\_Interface, Surfaces, Graphite\_Layer\_Sep \\
STRATUM\_4 & Defects & Defects \\
STRATUM\_5 & Finite/low-dimensional & Dimer, Fullerenes, Graphene, Nanotubes, LD\_Iter1, Single\_Atom \\
\bottomrule
\end{tabular}%
}

\vspace{0.7em}
\noindent\textit{Panel C: rMD17 evaluation strata.} Molecules are paired by increasing atom count, with lexicographic ordering for ties. Each molecule contributes 100/20/20 train/validation/test frames and 20/80 calibration/candidate frames within training.\\[2pt]
\begin{tabular}{lll}
\toprule
Code & Molecules & Atom counts \\
\midrule
STRATUM\_1 & ethanol, malonaldehyde & 9, 9 \\
STRATUM\_2 & benzene, uracil & 12, 12 \\
STRATUM\_3 & toluene, salicylic acid & 15, 16 \\
STRATUM\_4 & naphthalene, paracetamol & 18, 20 \\
STRATUM\_5 & aspirin, azobenzene & 21, 24 \\
\bottomrule
\end{tabular}

\vspace{0.7em}
\noindent\textit{Panel D: validation-only choice of the coverage comparator.}\\[2pt]
\begin{tabular}{lll}
\toprule
Dataset & Budget & Retained comparator \\
\midrule
Carbon & 1\%  & FPS \\
Carbon & 5\%  & DPP \\
Carbon & 20\% & DPP \\
rMD17 full & 1\%  & DPP \\
rMD17 full & 5\%  & DPP \\
rMD17 full & 20\% & DPP \\
\bottomrule
\end{tabular}
\end{table*}

\subsection{MACE outcome comparisons}
\label{sec:si_mace_protocol}

The primary Carbon and pooled rMD17 outcome comparisons use the official MACE implementation~\cite{mace}. The architecture, label conventions, loss weights, optimiser, batch sizes, learning-rate schedule, epoch ceiling and stopping settings described here are shared by every selector at 1\%, 5\% and 20\% and by the full-data references. Selected subsets contain 10, 50 or 200 structures, and the full-data reference uses the complete 1000-structure training partition. The 200-structure validation and 200-structure test partitions are shared across selectors and retained-data budgets. Outcome seeds 1--5 are paired between subset models and their full-data references.

All compression models used CUDA execution and float64 precision. The two interaction blocks were \texttt{RealAgnosticInteractionBlock} followed by \texttt{RealAgnosticResidualInteractionBlock}. The 5.0~\AA\ radial cutoff used eight Bessel functions and a polynomial envelope with $p=5$. The maximum spherical-harmonic degree was 3, correlation order was 3 and hidden irreducible representations were \texttt{128x0e + 128x1o}. The radial MLP had widths \texttt{[64,64,64]}, and the final readout used \texttt{16x0e} with a SiLU gate.

Reference labels were read from \texttt{dft\_energy} in eV and \texttt{dft\_forces} in eV~$\text{\AA}^{-1}$. The weighted loss used energy and force weights of 1 and 100, respectively, with unit weight for the Default configuration type. Force computation was enabled and stress computation disabled. Atomic reference energies were estimated with \texttt{E0s=average}; RMS-force output scaling and automatic mean-neighbour normalisation were estimated from each model's own training set. Their estimation procedures were fixed, but their fitted values could differ between selected subsets and the full-data reference.

AdamW used AMSGrad, an initial learning rate of 0.01 and weight-decay coefficient $5\times10^{-7}$ in the standard MACE parameter groups. Training and validation batch sizes were four, and gradient clipping was set to 10. The epoch ceiling was 1500, validation was evaluated every epoch, ReduceLROnPlateau used factor 0.8 and patience 50, and early-stopping patience was configured as 200 epochs. Exponential moving averaging used decay 0.99; stage-two/SWA training was disabled. Checkpoint retention was configured with \texttt{keep\_checkpoints=false} and \texttt{save\_all\_checkpoints=false}, automatic restart was disabled and CPU model export was enabled. The official report format was \texttt{PerAtomMAE}; the force-vector endpoints analysed here are defined separately below. The shared settings are specified by \path{training_configs/DD_Carbon_FPS_DPP_B005_s1.yaml} in the manuscript archive. Run names, training-file paths, output directories and training seeds vary by run; the validation-file path is fixed within each dataset.

The retained-data budget counts training structures. An epoch traverses the corresponding training set, so the same epoch ceiling does not impose the same gradient-update budget at different subset sizes. Learning-rate reduction and early stopping respond to validation loss within this shared schedule. The comparisons therefore evaluate selector performance under a common epoch-based training protocol.

For each seed, the same held-out atom sets are used to calculate the four force endpoints. Relative gaps are formed against that seed's full-data reference before averaging across seeds. This pairing also defines the selector contrasts in Supplementary Tables~\ref{tab:si_precision_summary} and~\ref{tab:si_20pct_paired}. The frozen MACE-MP-0 model used to compute selection descriptors is separate from the outcome models. The Carbon-defect foundation-model comparison uses fine-tuning and is reported separately.

\subsection{Coverage baselines and candidate signals}
\label{sec:si_selectors_criteria}

Random sampling is performed without replacement. FPS and DPP operate on standardised, mean-pooled invariant MACE-MP-0 descriptor embeddings. FPS begins with the structure farthest from the centroid. The DPP uses greedy maximum-a-posteriori selection from an RBF $L$-ensemble: its bandwidth is the median positive pair distance, diagonal jitter is $10^{-10}$, exact selection is used up to 4096 candidates, and larger pools use 256 random Fourier features with seed 0. Gaussian-kernel QUESTS coverage and its minimum-set-cover variant, QUESTS-MSC, use bandwidth 0.015, 32 neighbours and a 5~\AA\ cutoff. This fixed QUESTS-MSC parameterisation is not tuned by dataset or budget. The predeclared validation endpoint is normalised force MAE, with lower values preferred, tolerance $10^{-6}$ and FPS followed by lexicographic structure ID as the exact-tie rule. Validation retains FPS for Carbon at 1\%, DPP for Carbon at 5\% and 20\%, and DPP for rMD17 at all three budgets; Table S0 lists the retained comparator at each budget. FPS/DPP is used only as a compact cross-budget family label. QUESTS-MSC is a secondary coverage comparison and Random is the unstructured reference.

Candidate analyses comprise learned chemical/physical similarity, the HEA25 descriptor-conditioned non-spin energy residual, frozen MACE-MP-0 force error and response-based selection. The first two are diagnostics without retraining. Foundation-model error is evaluated through one Carbon-defect fine-tuning endpoint. Response effects are evaluated for the Metadata-score mask, High-response cell, High-force tail and High-novelty mask errors. The Metadata-score mask is the upper quintile of the calibration-defined composite physical-stress score given below; the High-novelty mask is the calibration-defined high-novelty atom subset. For candidate method $m$, the reported 5\% effect is $100[L_g(\theta_m)-L_g(\theta_{\mathrm{FPS/DPP}})]/L_g(\theta_{\mathrm{FPS/DPP}})$; positive values indicate higher error. All interpretations are made against the strongest available FPS/DPP baseline at the same budget.

\subsection{Definitions and procedures}
\label{sec:si_methods_detail}

This section defines the force endpoints, learned-metric objective, HEA regression target and response-selection procedure. It expands the three response coordinates in main-text Eq.~\ref{M-eq:response_coordinates} with the complete calibration and cell-construction procedure.

\noindent\textbf{Outcome endpoint and evaluation strata.}
For a subset $S$ and full pool $\mathcal{D}$, models $\theta_S$ and $\theta_{\mathcal{D}}$ are compared on common held-out structures using the pairing in Sec.~\ref{sec:si_mace_protocol}. Write $e_{xa}(\theta)=\lVert\mathbf F_{xa}^{\theta}-\mathbf F_{xa}^{\mathrm{DFT}}\rVert_2$ for the force-vector error of atom $a$ in held-out structure $x$. An evaluation group $g$ is a fixed, potentially overlapping atom set: overall, the High-force tail, the Metadata-score mask, the High-response cell, the Validation-selected stratum or the High-novelty mask. For the overall, High-force tail, Metadata-score, High-response-cell and Validation-selected-stratum endpoints, $L_g(\theta)=|g|^{-1}\sum_{(x,a)\in g}e_{xa}(\theta)$. For the High-novelty mask, $L_g(\theta)=\{\,|g|^{-1}\sum_{(x,a)\in g}e_{xa}^{2}(\theta)\,\}^{1/2}$. We report both the signed relative deviation and its magnitude,
\[
r_g(S)=100\,\frac{L_g(\theta_S)-L_g(\theta_{\mathcal D})}{\max\{L_g(\theta_{\mathcal D}),\epsilon\}},
\qquad \delta_g(S)=|r_g(S)|,
\]
with $\epsilon=10^{-12}$. The absolute gap $\delta_g$ measures proximity to the full-data outcome, while $r_g$ retains the direction of the error difference. The High-force tail mask is fixed from the 80th percentile of the DFT force norm in response calibration: 0.423620725~eV~$\text{\AA}^{-1}$ for Carbon and 0.324778407~eV~$\text{\AA}^{-1}$ for rMD17.

The Validation-selected stratum is chosen once from full-reference validation predictions,
\begin{equation}
g^\star=\arg\max_g\frac{1}{5}\sum_{s=1}^{5}L_{g,\mathrm{val}}(\theta_{\mathcal D,s}),
\label{eq:si_validation_group}
\end{equation}
with lexicographic tie-breaking. Carbon uses the five physical groups in Table S0 and selects finite/low-dimensional Carbon (STRATUM\_5). The pooled rMD17 analysis uses the five equal-size molecule-pair strata in Table S0 and selects benzene/uracil (STRATUM\_2). The selected group is frozen before test evaluation and is shared by all selectors, budgets and seeds.

The High-novelty mask is a high-novelty environment subset. Let $\mathbf h_{xa}$ be the invariant atomic embedding from the frozen MACE-MP-0 medium model and let $\mathcal H_{\mathcal K}$ be the response-calibration environment bank. We define
\begin{equation}
n_{xa}=d_{5\mathrm{NN}}(\mathbf h_{xa},\mathcal H_{\mathcal K}),\qquad
M^{\mathrm{nov}}_{xa}=\mathbb{I}\!\left[n_{xa}\ge Q_{0.65}^{\mathcal K}(n)\right].
\label{eq:si_novelty_mask}
\end{equation}
When the calibration novelty distribution is formed, atoms from the query structure are omitted from the reference bank. The fitted 65th-percentile threshold is projected unchanged to validation and test. The endpoint is the RMS force-vector error on $M^{\mathrm{nov}}$ and does not represent trajectory drift.

The Metadata-score mask combines three physical stress indicators without using a model's held-out error. Let $R_{\mathcal K}$ denote the calibration empirical-rank map, let $e_x=E_x^{\mathrm{DFT}}/N_x$, and let $\widetilde e_{g(x)}$ be the calibration median energy per atom in the Carbon physical group or rMD17 molecule of $x$. The atom-level score and mask are
\begin{equation}
\begin{aligned}
s_{xa}&=\frac{1}{3}\Bigl[R_{\mathcal K}(|e_x-\widetilde e_{g(x)}|)
+R_{\mathcal K}(\lVert\mathbf F_{xa}^{\mathrm{DFT}}\rVert_2)
+R_{\mathcal K}(n_{xa})\Bigr],\\
M^{\mathrm{stress}}_{xa}&=\mathbb{I}\!\left[s_{xa}\ge Q_{0.8}^{\mathcal K}(s)\right].
\end{aligned}
\label{eq:si_metadata_score}
\end{equation}
Every median, empirical-rank map and threshold is estimated on response calibration and then held fixed. Direct held-out force errors are additionally reported for the 20\% response crossover because the witness itself is defined against the full-data model.

\noindent\textbf{Learned-metric fit objective.}
The channel-weighted metric is fitted only on the declared training fit split. For structure $x$, define
\begin{equation}
\mathbf v_x=\left[\Delta e_x,\ \log(\overline{\lVert\mathbf F_x^{\mathrm{DFT}}\rVert}+\epsilon),\ \log\!\left(Q_{0.9}(\lVert\mathbf F_x^{\mathrm{DFT}}\rVert)+\epsilon\right)\right],
\label{eq:si_pair_vector}
\end{equation}
where $e_x=E_x^{\mathrm{DFT}}/N_x$. The energy is centred within the Carbon physical group, rMD17 molecule or HEA FCC/BCC--generation-class group. All three components are standardised using the current fit fold only. Unique unordered pairs are enumerated when their number is no greater than 100000; otherwise 100000 pairs are sampled without replacement with pair seed 0. For pair $(i,j)$,
\begin{equation}
y_{ij}=\exp\!\left[-\frac{\lVert\widetilde{\mathbf v}_i-\widetilde{\mathbf v}_j\rVert_2^2}{2\ell^2}\right],
\qquad
\ell=\operatorname{median}_{i<j,\ d_{ij}>0}d_{ij},
\label{eq:si_pair_target}
\end{equation}
where $d_{ij}=\lVert\widetilde{\mathbf v}_i-\widetilde{\mathbf v}_j\rVert_2$. In every leave-one-group-out analysis, pairs touching the held group are excluded and the component scaling and $\ell$ are refitted from the remaining groups.

The distance combines the three input channels associated with geometry and coverage, species and composition, and the residual channel. Let $\mathbf u_k(x)\in\mathbb R^{p_k}$ denote channel $k$ for structure $x$. Each feature is standardised using its fit-fold mean and population standard deviation; a standard deviation no greater than $10^{-12}$ is replaced by one. With the transformed channels denoted by $\widetilde{\mathbf u}_k$, the squared distance is
\begin{equation}
\begin{aligned}
d_{\lambda}^{2}(x,y)&=\sum_{k=1}^{3}\frac{\lambda_k}{p_k}\lVert\widetilde{\mathbf u}_k(x)-\widetilde{\mathbf u}_k(y)\rVert_2^2,\\
\lambda_k&\ge0,\qquad \sum_k\lambda_k=1.
\end{aligned}
\label{eq:si_channel_distance}
\end{equation}
Division by $p_k$ averages over features within a channel before the channels are weighted. The predicted pair similarity is $\widehat{y}_{ij}=\exp[-d_{\lambda}^{2}(x_i,x_j)/(2\sigma^{2})]$. The interface uses $\sigma=1$, no regularisation and a maximum of 500 optimisation iterations. Non-negative channel weights constrained to sum to one minimise the pairwise cross-entropy
\begin{equation}
\mathcal{L}_{\mathrm{pair}}=-\frac{1}{|\mathcal{P}_{\mathrm{fit}}|}\sum_{(i,j)\in\mathcal{P}_{\mathrm{fit}}}\left[y_{ij}\log \widehat{y}_{ij}+(1-y_{ij})\log(1-\widehat{y}_{ij})\right].
\label{eq:si_learned_metric_fit}
\end{equation}
Neither leave-one-group-out AUROC nor any validation or held-out-test outcome endpoint enters the fit. The pair seed, fit IDs, sampled pair indices, fold-specific scaling and kernel scale are fixed before evaluation.

Ranking uses the negative logarithm of the mean kernel similarity to a fixed reference bank $\mathcal R$:
\begin{equation}
q_{\lambda}(x)=-\log\!\left[\frac{1}{|\mathcal R|}\sum_{y\in\mathcal R}\exp\!\left(-\frac{d_{\lambda}^{2}(x,y)}{2\sigma^2}\right)\right].
\label{eq:si_metric_novelty}
\end{equation}
Higher scores are selected first. When query and reference sets coincide, the self term is omitted and the denominator is $|\mathcal R|-1$. In leave-one-group-out evaluation the held group is excluded from the reference bank. The dimension normalisation, channel standardisation and kernel transformation are shared by pair fitting and subsequent scoring.

\noindent\textbf{Fixed-bank similarity baseline.}
The fixed-bank baseline is evaluated independently of the learned metric. In each leave-one-group-out fold, the descriptor environments from all non-held-out structures form a frozen reference bank. For a query environment $q$, we compute the maximum Gaussian similarity to that bank and define its novelty as one minus this maximum similarity; structure-level novelty is the mean over its environments. The held-out group contributes neither reference environments nor fitted parameters. The descriptor definition, bank membership and Gaussian bandwidth are fixed for every fold.

\noindent\textbf{Descriptor-strength ladder.}
The public HEA25 calculations are non-spin-polarised~\cite{hea}; the diagnostic target is therefore a non-spin DFT energy residual. Within each analysis fit fold, elemental reference coefficients $\epsilon_Z$ are obtained by least squares and
\begin{equation}
y_x=\frac{E_x^{\mathrm{DFT}}}{N_x}-\sum_Z c_{xZ}\epsilon_Z,
\label{eq:si_hea_target}
\end{equation}
where $c_{xZ}$ is the atomic fraction of element $Z$. The nested descriptor ladder is: $G_0$, the 25 elemental fractions, atom count, volume per atom, FCC/BCC indicator and generation-class indicators; $G_1$, $G_0$ plus a mean-pooled two-body radial spectrum with 6~\AA\ cutoff, Gaussian width 0.25~\AA\ and 12 radial functions; $G_2$, $G_1$ plus the corresponding rotationally invariant three-body angular power spectrum through $\ell=6$~\cite{soap}; $G_3$, $G_2$ plus mean-pooled ACE invariants through correlation order 3 (four-body) and $\ell_{\max}=3$~\cite{ace}; and $G_4$, $G_3$ plus the mean-pooled invariant output of the second interaction block of the frozen MACE-MP-0 medium model~\cite{macemp}.

Every feature is standardised on the fit fold. Ridge regression is fitted at each level, with its penalty selected from $10^{-6},10^{-5},\ldots,10^{3}$ by inner grouped cross-validation. Outer evaluation leaves out one of the eight FCC/BCC--generation-class groups. For residuals $r_x^{(k)}=y_x-\widehat f_k(G_k(x))$, the normalised out-of-fold squared error is
\begin{equation}
V_k=\frac{\sum_{x\in\mathrm{OOF}}(r_x^{(k)})^2}{\sum_{x\in\mathrm{OOF}}(r_x^{(0)})^2}.
\label{eq:si_hea_floor}
\end{equation}
Thus $V_k$ compares the summed out-of-fold squared errors at level $k$ with those at G0. The comparison measures how the structural representation affects prediction of the non-spin DFT energy target.

\noindent\textbf{Coverage-conditioned residualisation.}
Let $u$ be the candidate score and $n$ the coverage novelty, defined as the rank-normalised $k$-nearest-neighbour distance in the working representation. We fit $u = h(n,\,\mathbf{z}) + u_\perp$, where $\mathbf{z}$ contains force scale, density or composition and local structural descriptors, and retain $u_\perp$. We report $R^2_{\mathrm{novelty}}$, $R^2_{\mathrm{nov+nuis}}$ and the partial correlation $\rho(u_\perp,n)$. Downstream value is not inferred from this diagnostic alone.

\noindent\textbf{Foundation-error residualisation.}
For the foundation analysis, the score is frozen MACE-MP-0 per-structure force error $u_F$. Novelty $n_F$ is the rank-normalised five-nearest-neighbour distance in the mean-pooled MACE-MP-0 embedding. Five-fold out-of-fold residualisation removes novelty and the nuisance variables. Raw-error and residual-error subsets are then compared with coverage and Random at the same size through held-out Carbon-defect force-vector MAE.

\noindent\textbf{Response witness, OBNL cells and repair.}
For budget $B$, let $S_{0,B}$ be the validation-selected FPS/DPP subset. A deterministic MACE seed-0 model trained on $S_{0,B}$ and a seed-0 full-data model trained on $\mathcal D$ construct the response selection. Outcome models are then trained independently with paired seeds 1--5. Write $\Delta\mathbf F_{xa,B}=\mathbf F_{\theta_{S_{0,B},0}}(x)_a-\mathbf F_{\theta_{\mathcal D,0}}(x)_a$. The structure witness is
\begin{equation}
W_B(x)=\frac{1}{N_x}\sum_a\lVert\Delta\mathbf F_{xa,B}\rVert_2.
\label{eq:si_response_witness}
\end{equation}
The response coordinates are
\begin{align}
q_E(x)&=z_{\mathcal K}\!\left(\frac{|E_{\theta_{S_{0,B},0}}(x)-E_{\theta_{\mathcal D,0}}(x)|}{N_x}\right),\\
q_{F,\mathrm{RMS}}(x)&=z_{\mathcal K}\!\left(\sqrt{\frac{1}{N_x}\sum_a\lVert\Delta\mathbf F_{xa,B}\rVert_2^2}\right),\\
q_{F,\max}(x)&=z_{\mathcal K}\!\left(\max_a\lVert\Delta\mathbf F_{xa,B}\rVert_2\right),
\label{eq:si_response_coordinates}
\end{align}
where each $z_{\mathcal K}(v)=(v-\mu_{\mathcal K})/\sigma_{\mathcal K}$ uses its own mean and population standard deviation from the fixed 200-structure response-calibration partition $\mathcal K$. If $\sigma_{\mathcal K}=0$, that coordinate is set to zero for calibration and candidates. No validation or test value enters the transformation.

Fixed edges $(-\infty,-1,0,1,+\infty)$ on each coordinate form a $4\times4\times4$ grid. The fitted transforms and edges assign both calibration structures and the disjoint 800 selection candidates to cells. A cell is eligible if it contains at least ten calibration structures. Its response priority is
\begin{equation}
R_B(c)=Q_{0.9}\{W_B(x):x\in\mathcal K,\ c(x)=c\}.
\label{eq:si_cell_priority}
\end{equation}
An eligible cell is high response when $R_B(c)$ is at or above the 90th percentile of priorities across eligible cells. Its candidate-pool occurrence mass is $O(c)=\sum_{x\in\mathcal P}N_x\mathbb{I}[c(x)=c]$, and it is occurred when $O(c)$ is at or above the median across eligible cells. A cell is OBNL when it is both high response and occurred but $S_{0,B}$ contains no member of that cell. Candidate-pool OBNL mass is $\sum_{x\in\mathcal P}N_x\mathbb{I}[c(x)\in\mathcal C_{\mathrm{OBNL}}]/\sum_{x\in\mathcal P}N_x$.

Global response-witness selection ranks candidates by $R_B(c(x))$ and breaks cell-level ties by lexicographic structure ID. Response repair replaces a specified fraction of $S_{0,B}$ with candidates from OBNL cells while preserving $B$; replacement fractions of 2\%, 5\% and 10\% are rounded upward. The primary 5\% setting replaces three of 50 Carbon structures, realising 6.0\%. In rMD17, coverage spans every high-response cell at 5\%, so no replacement is made. The frozen selection is trained with seeds 1--5 and evaluated only on the held-out test set. Because its construction requires a full-data model, the witness is a retrospective compression diagnostic rather than a pre-labelling acquisition rule.

\noindent\textbf{Statistical analysis.}
Paired training-seed contrasts use outcome seeds 1--5. Relative gaps are calculated within each seed against the matched full-data reference and then averaged. For Random minus FPS/DPP and witness minus FPS/DPP absolute gaps at 5\%, all $5^5$ with-replacement resamples of each five-value paired-difference vector are enumerated. The 2.5th and 97.5th percentiles use linear interpolation. At 20\%, the same procedure is applied to witness--FPS/DPP absolute-gap and direct-error differences and to the witness signed gap from the full-data reference. Paired $t$ intervals use four degrees of freedom. These intervals describe training-seed variability conditional on the selected subsets.

Grouped intervals use 10,000 percentile-bootstrap resamples of the relevant scientific groups: five Carbon physical groups, ten rMD17 molecules or eight HEA25 FCC/BCC--generation-class groups. The constructed learned-metric positive control uses its separately stated 2,000-resample protocol. The 1\% comparison reports five-seed means. Supplementary Table~\ref{tab:si_interval_map} identifies the uncertainty estimate for each contrast. CI half-widths are $(\mathrm{CI}_{\mathrm{high}}-\mathrm{CI}_{\mathrm{low}})/2$.

\section{Extended results}

\subsection{Static physical proxy diagnostics}
\label{sec:si_static_proxies}

Two descriptor diagnostics complement the retraining comparisons. Learned chemical/physical similarity is compared with Gaussian QUESTS through ranking and leave-one-group-out evaluation (main-text Sec.~\ref{M-sec:results_learned_similarity}). Its fitting objective is Eq.~\ref{eq:si_learned_metric_fit}; AUROC is evaluated after fitting. The auxiliary HEA analysis measures the normalised out-of-fold squared error of the non-spin DFT energy target across G0--G4. These diagnostics concern ranking distinctness and descriptor adequacy, respectively.

\begin{figure*}[!tp]
\centering
\includegraphics[width=172mm]{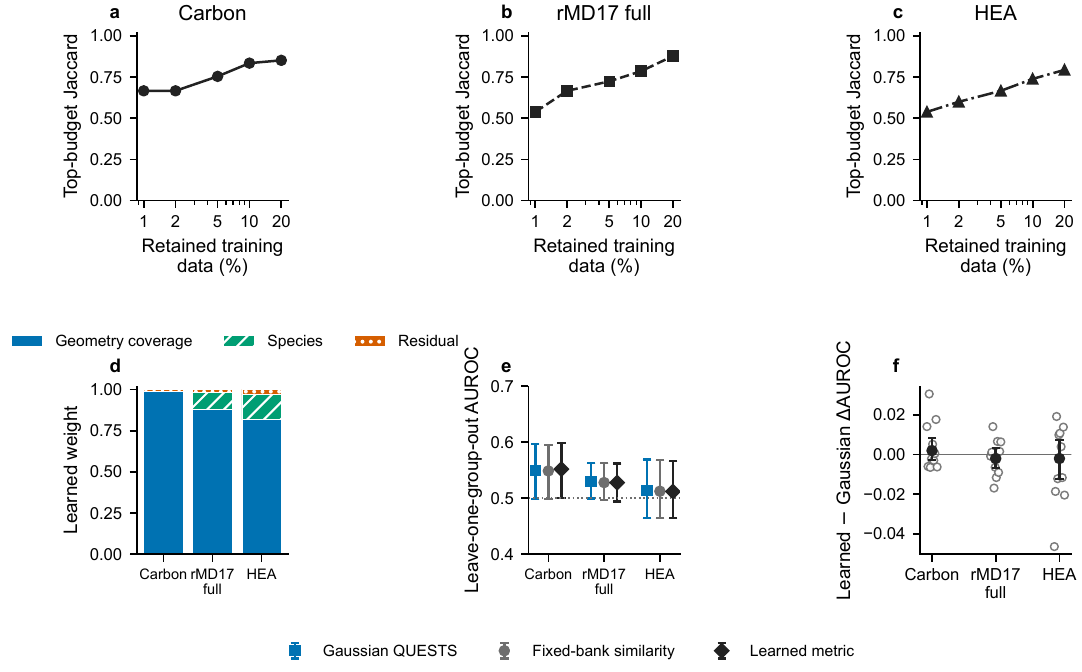}
\caption{\textbf{Learned chemical/physical similarity remains aligned with Gaussian QUESTS coverage.} \textbf{a--c}, Top-budget Jaccard overlap from 1\% to 20\% for Carbon, rMD17 full and HEA, respectively. \textbf{d}, Learned channel weights across datasets. \textbf{e}, Leave-one-group-out AUROC for Gaussian QUESTS, fixed-bank similarity and the learned metric. \textbf{f}, Held-out-group learned-minus-Gaussian $\Delta$AUROC values; black markers show grouped-bootstrap means and 95\% confidence intervals.}
\label{fig:si_learned_similarity}
\end{figure*}

The learned metric preserves much of the Gaussian QUESTS ranking (Supplementary Fig.~\ref{fig:si_learned_similarity}a--c). Spearman $\rho_S$ is 0.990 for Carbon, 0.990 for rMD17 full and 0.980 for HEA, with corresponding top-5\% Jaccard overlaps of 0.754, 0.724 and 0.667. Overlap increases across the 1--20\% range: from 0.667 to 0.852 for Carbon, 0.538 to 0.878 for rMD17 full and 0.538 to 0.794 for HEA. Leave-one-group-out AUROCs remain close to the Gaussian QUESTS and fixed-bank baselines; the learned-minus-Gaussian $\Delta$AUROC interval includes zero in each dataset.

\begin{table*}[!tp]
\centering
\small
\caption{\textbf{Summary statistics for learned similarity.} Panel A reports rank consistency, top-budget overlap and leave-one-group-out AUROC. Panel B reports fitted channel weights. $\Delta$AUROC is the learned metric minus Gaussian QUESTS; brackets show grouped-bootstrap 95\% confidence intervals.}
\label{tab:si_learned_similarity_summary}

\adjustbox{max width=\textwidth}{%
\begin{tabular}{lcccccccccc}
\toprule
Dataset
& $\rho_S$
& $J_{1\%}$
& $J_{2\%}$
& $J_{5\%}$
& $J_{10\%}$
& $J_{20\%}$
& AUROC$_{\mathrm{Gauss}}$
& AUROC$_{\mathrm{fixed}}$
& AUROC$_{\mathrm{learned}}$
& $\Delta$AUROC \\
\midrule
Carbon
& $0.990$
& $0.667$
& $0.667$
& $0.754$
& $0.835$
& $0.852$
& $0.550$
& $0.549$
& $0.552$
& $+0.002$ [$-0.003$, $+0.008$] \\
rMD17 full
& $0.990$
& $0.538$
& $0.667$
& $0.724$
& $0.786$
& $0.878$
& $0.530$
& $0.528$
& $0.528$
& $-0.002$ [$-0.007$, $+0.003$] \\
HEA
& $0.980$
& $0.538$
& $0.600$
& $0.667$
& $0.739$
& $0.794$
& $0.514$
& $0.513$
& $0.512$
& $-0.002$ [$-0.013$, $+0.007$] \\
\bottomrule
\end{tabular}%
}

\vspace{0.8em}

\adjustbox{max width=\textwidth}{%
\begin{tabular}{lccc}
\toprule
Dataset
& $w_{\mathrm{geom}}$
& $w_{\mathrm{species}}$
& $w_{\mathrm{res}}$ \\
\midrule
Carbon
& $0.989$
& n/a
& $0.011$ \\
rMD17 full
& $0.880$
& $0.101$
& $0.020$ \\
HEA
& $0.821$
& $0.149$
& $0.030$ \\
\bottomrule
\end{tabular}%
}
\end{table*}

Geometry and coverage dominate the fitted channel weights (Supplementary Table~\ref{tab:si_learned_similarity_summary}); the residual channel receives at most 0.030. Together with $\rho_S\geq0.980$, the increasing top-budget overlap and the near-zero AUROC differences, these weights describe a metric that remains closely aligned with coverage under the evaluated parameterisation. This ranking result is distinct from the retrained outcomes reported for the response and foundation-error selectors.

\begin{table*}[!tp]
\centering
\small
\caption{\textbf{Normalised out-of-fold squared error for non-spin HEA energy prediction.} The regression target and error normalisation are defined in Eqs.~\ref{eq:si_hea_target} and~\ref{eq:si_hea_floor}, and the cumulative G0--G4 feature sets are specified in the descriptor-strength ladder. Values are out-of-fold residual sums of squares normalised to G0; intervals are 95\% grouped-bootstrap intervals over the eight FCC/BCC--generation-class groups.}
\label{tab:si_residual_gap}
\begin{tabular}{lll}
\toprule
Level & Cumulative representation & Normalised squared error (95\% CI) \\
\midrule
G0 & Composition, cell and generation controls & $1.000$ $[1.000,\,1.000]$ \\
G1 & G0 + two-body radial spectrum & $0.720$ $[0.669,\,0.805]$ \\
G2 & G1 + three-body angular spectrum & $0.500$ $[0.466,\,0.560]$ \\
G3 & G2 + ACE invariants & $0.320$ $[0.290,\,0.371]$ \\
G4 & G3 + frozen MACE embedding & $0.240$ $[0.222,\,0.278]$ \\
\bottomrule
\end{tabular}
\end{table*}

The normalised out-of-fold squared error decreases at each step of the descriptor ladder (Supplementary Table~\ref{tab:si_residual_gap}). G4 reaches 0.240 of the G0 value, with a grouped-bootstrap interval of $[0.222,0.278]$. The richer representation therefore predicts the non-spin DFT energy target more accurately under grouped cross-validation. This comparison measures descriptor adequacy rather than the downstream value of a selected training subset.

\subsection{Foundation-model error analysis}
\label{sec:si_foundation_error}

\begin{figure*}[!tp]
\centering
\includegraphics[width=172mm]{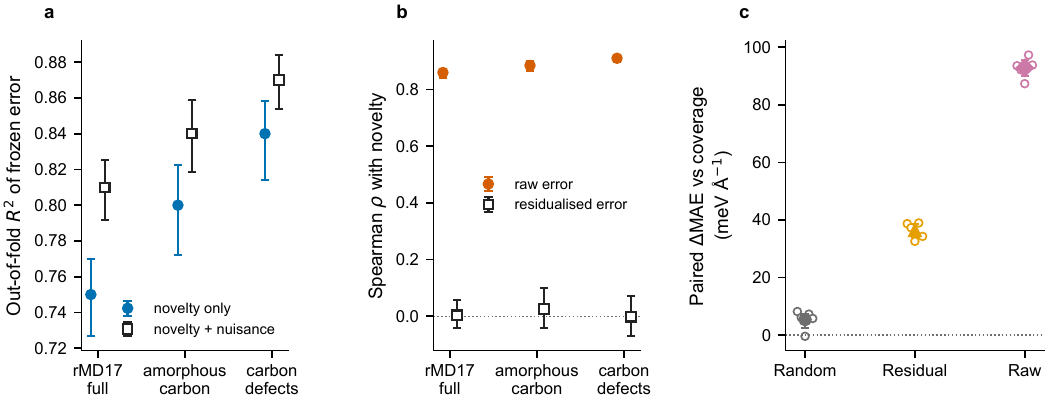}
\caption{\textbf{Foundation-model error analysis.} \textbf{a}, Out-of-fold variance explained by embedding novelty alone and by novelty plus nuisance covariates across rMD17, amorphous Carbon and Carbon defects. \textbf{b}, Raw and residualised error correlations with novelty. In \textbf{a} and \textbf{b}, points and error bars show the reported estimates and 95\% confidence intervals. \textbf{c}, Carbon-defect force-MAE differences from embedding coverage: open points show all five paired training seeds and filled points with error bars show the reported aggregate differences and 95\% confidence intervals. Force MAE is in meV~$\text{\AA}^{-1}$.}
\label{fig:si_foundation_error}
\end{figure*}

We evaluate frozen-model error as a selection signal in two steps (Sec.~\ref{M-sec:results_foundation_error}). First, we relate MACE-MP force error to embedding novelty and nuisance covariates. We then compare raw-error and residual-error selections through Carbon-defect fine-tuning.

Frozen-model error tracks embedding novelty in rMD17 full, amorphous Carbon and Carbon defects. Their raw error--novelty correlations are 0.860, 0.885 and 0.911, respectively, and novelty alone explains 0.75, 0.80 and 0.84 of the out-of-fold error variance. Residualisation reduces the corresponding correlations to 0.004, 0.026 and $-0.002$.

Across seeds 1--5, mean held-out force MAEs are 226.62 for Random, 221.22 for Embedding coverage, 257.61 for Residualised error and 314.06~meV~$\text{\AA}^{-1}$ for Raw error. Relative to embedding coverage, the paired differences are $+5.40$, $+36.39$ and $+92.84$~meV~$\text{\AA}^{-1}$ for Random, Residualised error and Raw error, respectively.

\begin{table*}[!tp]
\centering
\scriptsize
\caption{\textbf{Foundation-model error statistics.} Panel A reports the decomposition across rMD17, amorphous Carbon and Carbon defects. Panels B and C report the Carbon-defect five paired seeds and grouped-bootstrap contrasts at the 5\% budget.}
\label{tab:si_foundation_error_summary}
\noindent\textit{Panel A: error decomposition and residualisation.}\\[2pt]
\resizebox{\textwidth}{!}{%
\begin{tabular}{llll}
\toprule
Dataset & Metric & Estimate & 95\% CI \\
\midrule
rMD17 & Novelty-only out-of-fold $R^2$ & $0.750$ & $[0.727,\,0.770]$ \\
rMD17 & Novelty+nuisance out-of-fold $R^2$ & $0.810$ & $[0.792,\,0.825]$ \\
rMD17 & Raw error--novelty Spearman $\rho$ & $0.860$ & $[0.842,\,0.873]$ \\
rMD17 & Residual--novelty Spearman $\rho$ & $0.004$ & $[-0.043,\,0.057]$ \\
Amorphous Carbon & Novelty-only out-of-fold $R^2$ & $0.800$ & $[0.772,\,0.822]$ \\
Amorphous Carbon & Novelty+nuisance out-of-fold $R^2$ & $0.840$ & $[0.819,\,0.859]$ \\
Amorphous Carbon & Raw error--novelty Spearman $\rho$ & $0.885$ & $[0.866,\,0.901]$ \\
Amorphous Carbon & Residual--novelty Spearman $\rho$ & $0.026$ & $[-0.041,\,0.099]$ \\
Carbon defects & Novelty-only out-of-fold $R^2$ & $0.840$ & $[0.814,\,0.858]$ \\
Carbon defects & Novelty+nuisance out-of-fold $R^2$ & $0.870$ & $[0.854,\,0.884]$ \\
Carbon defects & Raw error--novelty Spearman $\rho$ & $0.911$ & $[0.897,\,0.922]$ \\
Carbon defects & Residual--novelty Spearman $\rho$ & $-0.002$ & $[-0.070,\,0.072]$ \\
\bottomrule
\end{tabular}%
}
\vspace{6pt}
\noindent\textit{Panel B: Carbon-defect five-seed held-out force MAE (meV~$\text{\AA}^{-1}$).}\\[2pt]
\resizebox{\textwidth}{!}{%
\begin{tabular}{llllllll}
\toprule
Selector & Seed 1 & Seed 2 & Seed 3 & Seed 4 & Seed 5 & Mean & 95\% CI \\
\midrule
Random & $218.94$ & $224.71$ & $227.83$ & $232.46$ & $229.18$ & $226.62$ & $[222.40,\,230.25]$ \\
Embedding coverage & $210.74$ & $218.61$ & $228.23$ & $225.16$ & $223.38$ & $221.22$ & $[215.46,\,226.03]$ \\
Residualised error & $249.47$ & $255.94$ & $260.82$ & $264.13$ & $257.69$ & $257.61$ & $[253.35,\,261.55]$ \\
Raw error & $304.31$ & $310.76$ & $315.58$ & $322.47$ & $317.19$ & $314.06$ & $[308.18,\,319.07]$ \\
\bottomrule
\end{tabular}%
}
\vspace{6pt}
\noindent\textit{Panel C: Carbon-defect grouped-bootstrap paired differences (meV~$\text{\AA}^{-1}$).}\\[2pt]
\resizebox{\textwidth}{!}{%
\begin{tabular}{lll}
\toprule
Comparison & $\Delta$MAE & 95\% CI \\
\midrule
Random minus embedding coverage & $+5.40$ & $[+2.44,\,+7.48]$ \\
Residualised error minus embedding coverage & $+36.39$ & $[+34.16,\,+38.55]$ \\
Raw error minus embedding coverage & $+92.84$ & $[+89.89,\,+95.53]$ \\
\bottomrule
\end{tabular}%
}
\end{table*}

Removing the rank association with novelty does not improve the error-based selector in the Carbon-defect comparison (Supplementary Table~\ref{tab:si_foundation_error_summary}). Embedding coverage gives the lowest mean held-out force MAE. Random, residual-error and raw-error selection are higher by 5.40, 36.39 and 92.84~meV~$\text{\AA}^{-1}$, respectively, with all three paired intervals above zero. The residual score therefore changes which error variation is selected without improving this fine-tuning endpoint.

\subsection{Response witness and repair}
\label{sec:si_response_repair}

Response witness and repair test the consequences of selecting high-response regions (Sec.~\ref{M-sec:results_response_repair}). At each budget, validation selects FPS or DPP to define $S_{0,B}$. The seed-0 coverage and full-data models then provide $W_B$ and the three response coordinates in Eqs.~\ref{eq:si_response_witness} and~\ref{eq:si_response_coordinates}. Their standardisation, 64-cell grid and priority $R_B(c)$ (Eq.~\ref{eq:si_cell_priority}) are fixed on calibration and projected to the disjoint candidate pool $\mathcal P$. An OBNL cell has at least ten calibration structures, upper-decile response priority, candidate-pool atom mass at or above the eligible-cell median, and no member in $S_{0,B}$. Its reported mass is $\eta_{\mathrm{OBNL}}=\sum_{x\in\mathcal{P}}N_x\mathbb{I}[c(x)\in\mathcal{C}_{\mathrm{OBNL}}]/\sum_{x\in\mathcal{P}}N_x$. Witness selection ranks candidates by $R_B(c)$, whereas repair replaces a fixed fraction of $S_{0,B}$ with candidates from $\mathcal C_{\mathrm{OBNL}}$. Both selections are constructed without validation or test structures and retrained with paired seeds 1--5 for held-out evaluation.

\begin{figure*}[!tp]
\centering
\includegraphics[width=172mm]{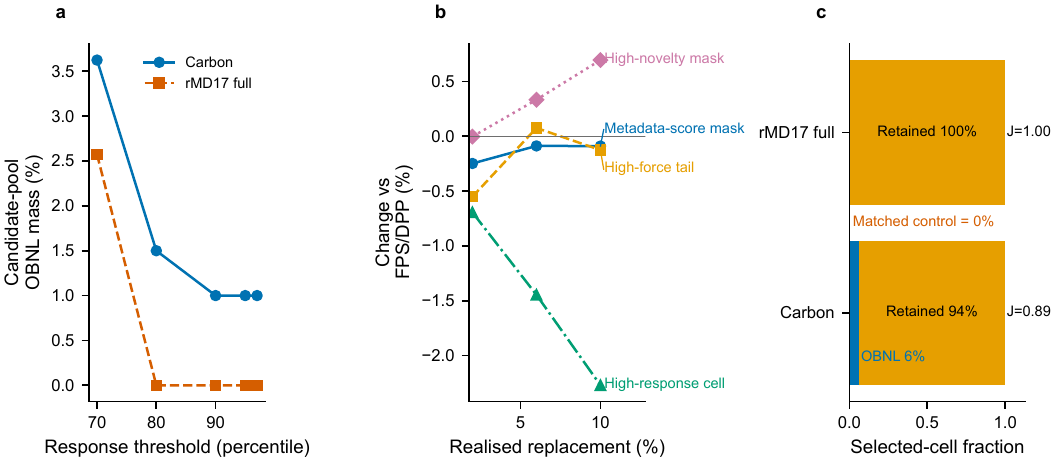}
\caption{\textbf{Response-cell and repair sensitivity.} \textbf{a}, Candidate-pool OBNL atom mass as the response threshold varies for Carbon and rMD17. \textbf{b}, Five-seed mean Carbon effects relative to FPS/DPP across realised replacement fractions of 2\%, 6\% and 10\%; curves are labelled by endpoint. \textbf{c}, Selected-cell composition and repair-versus-coverage Jaccard overlap at the primary 5\% budget. Carbon retains 94\% of the coverage subset and replaces 6\% with OBNL-cell representatives; rMD17 retains 100\%. Cells and thresholds are fitted on training calibration and projected to the disjoint candidate pool. Panels show point estimates from the sensitivity analysis.}
\label{fig:si_response_repair}
\end{figure*}

At 5\%, coverage leaves 1.00\% of the Carbon candidate-pool atom mass in OBNL cells, compared with 0.00\% for rMD17 (Fig.~\ref{fig:si_response_repair}; Table~\ref{tab:si_response_repair_summary}). The repair-versus-coverage Jaccard overlaps are 0.887 and 1.000, respectively. Carbon witness selection raises all four risks by $+6.33\%$ to $+59.45\%$, while all four repair intervals include zero. In rMD17, witness selection lowers High-response cell risk by 7.87\% but raises the other three risks. FPS/DPP already covers every high-response cell at this budget, so rMD17 repair makes no replacement and has zero effect at every endpoint.

\begin{table*}[!tp]
\centering
\scriptsize
\caption{\textbf{Response-witness and repair summary at the 5\% budget.} Panel A reports candidate-pool OBNL mass and fixed-selection overlap. Panel B reports effects relative to FPS/DPP; positive values indicate higher risk. Carbon replaces 3/50 structures, while rMD17 makes no replacement.}
\label{tab:si_response_repair_summary}
\noindent\textit{Panel A: frozen-selection occurrence and overlap summaries.}\\[2pt]
\resizebox{\textwidth}{!}{%
\begin{tabular}{lllll}
\toprule
System & Quantity & Estimate & Uncertainty interval & Definition \\
\midrule
Carbon & repair Jaccard vs FPS/DPP & $0.887$ & Not applicable (fixed value) & fixed-budget set overlap \\
Carbon & candidate-pool OBNL mass (\%) & $1.000$ & Not applicable (fixed value) & candidate-pool atom-weighted mass \\
rMD17 full & repair Jaccard vs FPS/DPP & $1.000$ & Not applicable (fixed value) & fixed-budget set overlap \\
rMD17 full & candidate-pool OBNL mass (\%) & $0.000$ & Not applicable (fixed value) & candidate-pool atom-weighted mass \\
\bottomrule
\end{tabular}%
}
\vspace{6pt}
\noindent\textit{Panel B: response effects at the primary 5\% budget.}\\[2pt]
\resizebox{\textwidth}{!}{%
\begin{tabular}{llllll}
\toprule
Dataset & Method & Endpoint & Effect & 95\% CI & Interpretation \\
\midrule
Carbon & Repair & Metadata-score mask & $-0.088\%$ & $[-0.365\%,\,+0.178\%]$ & CI includes zero \\
Carbon & Repair & High-force tail & $+0.076\%$ & $[-0.395\%,\,+0.570\%]$ & CI includes zero \\
Carbon & Repair & High-response cell & $-1.446\%$ & $[-3.484\%,\,+0.672\%]$ & CI includes zero \\
Carbon & Repair & High-novelty mask & $+0.333\%$ & $[-0.049\%,\,+0.724\%]$ & CI includes zero \\
Carbon & Witness & Metadata-score mask & $+59.446\%$ & $[+49.127\%,\,+71.695\%]$ & Higher risk \\
Carbon & Witness & High-force tail & $+40.471\%$ & $[+33.509\%,\,+48.234\%]$ & Higher risk \\
Carbon & Witness & High-response cell & $+6.329\%$ & $[+2.185\%,\,+13.521\%]$ & Higher risk \\
Carbon & Witness & High-novelty mask & $+46.095\%$ & $[+38.435\%,\,+54.332\%]$ & Higher risk \\
rMD17 & Repair & Metadata-score mask & $0.000\%$ & $[0.000\%,\,0.000\%]$ & Identical subset \\
rMD17 & Repair & High-force tail & $0.000\%$ & $[0.000\%,\,0.000\%]$ & Identical subset \\
rMD17 & Repair & High-response cell & $0.000\%$ & $[0.000\%,\,0.000\%]$ & Identical subset \\
rMD17 & Repair & High-novelty mask & $0.000\%$ & $[0.000\%,\,0.000\%]$ & Identical subset \\
rMD17 & Witness & Metadata-score mask & $+58.515\%$ & $[+50.864\%,\,+67.402\%]$ & Higher risk \\
rMD17 & Witness & High-force tail & $+13.200\%$ & $[+5.481\%,\,+21.187\%]$ & Higher risk \\
rMD17 & Witness & High-response cell & $-7.872\%$ & $[-16.547\%,\,-1.607\%]$ & Lower risk \\
rMD17 & Witness & High-novelty mask & $+16.971\%$ & $[+12.175\%,\,+21.919\%]$ & Higher risk \\
\bottomrule
\end{tabular}%
}
\end{table*}

\subsection{Outcome-gap evaluation}
\label{sec:si_outcome_gap}

We separate the outcome-gap comparison in Sec.~\ref{M-sec:results_budget_ordering} into training-seed and evaluation-stratum results. For Carbon and rMD17, we report Mean, High-force tail and High-novelty mask gaps alongside the distribution across the five strata. All percentage gaps use $\delta_g(S)=100|L_g(\theta_S)-L_g(\theta_{\mathcal D})|/\max\{L_g(\theta_{\mathcal D}),\epsilon\}$. Lower values indicate closer agreement with the full-data outcome; the signed DFT-risk difference is assessed separately.

\begin{figure*}[!tp]
\centering
\includegraphics[width=172mm]{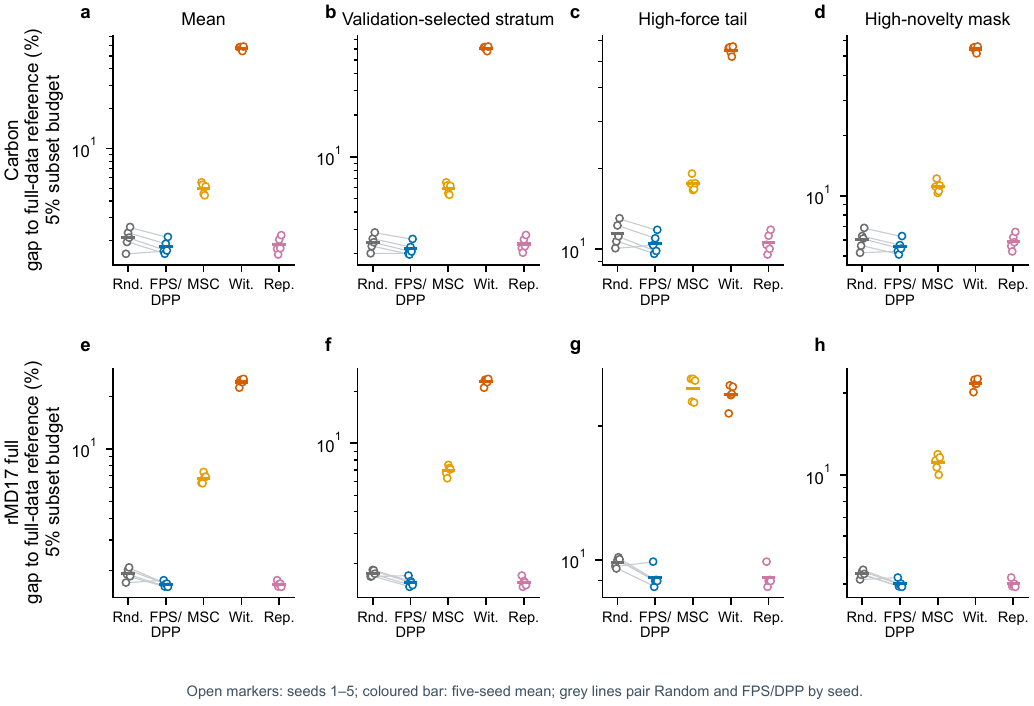}
\caption{\textbf{Outcome gaps for seeds 1--5 at the 5\% budget.} Carbon (\textbf{a--d}) and rMD17 (\textbf{e--h}) panels show seed-level values and five-seed means for Random, FPS/DPP, QUESTS-MSC, response witness and response repair across the Mean, Validation-selected stratum, High-force tail and High-novelty mask endpoints. Grey lines join Random and FPS/DPP values from the same training seed. Lower absolute gaps indicate closer agreement with the full-data training outcome.}
\label{fig:si_outcome_gap}
\end{figure*}

Across seeds 1--5, the 5\% FPS/DPP mean gaps are 1.80\% for Carbon and 1.68\% for rMD17. Random selection gives mean gaps of 2.10\% and 1.92\%, respectively, whereas QUESTS-MSC and response-witness selection deviate more strongly.

The validation-selected stratum gaps for FPS/DPP are 2.17\% for Carbon and 1.56\% for rMD17; the corresponding high-force-tail gaps are 10.52\% and 9.11\%.

At 5\%, FPS/DPP High-novelty mask gaps are 5.55\% for Carbon and 4.01\% for rMD17. The rMD17 repair reproduces the FPS/DPP subset and therefore has identical outcomes.

\begin{table*}[!tp]
\centering
\scriptsize
\renewcommand{\arraystretch}{0.94}
\caption{\textbf{Outcome-gap statistics across five seeds and evaluation strata at the 5\% budget.} All gaps are absolute percentages relative to the corresponding full-data training outcome. Panel B first averages each coded stratum over seeds 1--5, then reports the median and IQR across exactly STRATUM\_1--STRATUM\_5; it excludes the Mean, High-force tail and High-novelty mask endpoints. Its final column is the validation-selected coded stratum.}
\label{tab:si_outcome_gap}
\noindent\textit{Panel A: seed-level mean gap to full.}\\[2pt]
\resizebox{\textwidth}{!}{%
\begin{tabular}{llllllll}
\toprule
Dataset & Selector & Seed 1 & Seed 2 & Seed 3 & Seed 4 & Seed 5 & Mean \\
\midrule
Carbon & Random & $1.59\%$ & $1.96\%$ & $2.30\%$ & $2.11\%$ & $2.54\%$ & $2.10\%$ \\
Carbon & FPS/DPP & $1.67\%$ & $1.59\%$ & $1.90\%$ & $1.69\%$ & $2.14\%$ & $1.80\%$ \\
Carbon & QUESTS-MSC & $5.54\%$ & $5.31\%$ & $4.53\%$ & $4.41\%$ & $5.15\%$ & $4.99\%$ \\
Carbon & Response witness & $58.12\%$ & $58.46\%$ & $56.30\%$ & $54.45\%$ & $59.03\%$ & $57.27\%$ \\
Carbon & Response repair & $1.74\%$ & $1.57\%$ & $2.04\%$ & $1.76\%$ & $2.20\%$ & $1.86\%$ \\
rMD17 full & Random & $1.71\%$ & $1.91\%$ & $2.02\%$ & $2.09\%$ & $1.86\%$ & $1.92\%$ \\
rMD17 full & FPS/DPP & $1.77\%$ & $1.63\%$ & $1.68\%$ & $1.69\%$ & $1.62\%$ & $1.68\%$ \\
rMD17 full & QUESTS-MSC & $6.37\%$ & $6.35\%$ & $7.37\%$ & $6.90\%$ & $6.92\%$ & $6.78\%$ \\
rMD17 full & Response witness & $22.40\%$ & $24.76\%$ & $24.23\%$ & $24.12\%$ & $25.16\%$ & $24.13\%$ \\
rMD17 full & Response repair & $1.77\%$ & $1.63\%$ & $1.68\%$ & $1.69\%$ & $1.62\%$ & $1.68\%$ \\
\bottomrule
\end{tabular}%
}
\vspace{1pt}
\noindent\textit{Panel B: summary across evaluation strata.}\\[2pt]
\resizebox{\textwidth}{!}{%
\begin{tabular}{lllll}
\toprule
Dataset & Selector & Median across strata & IQR across strata & Validation-selected stratum \\
\midrule
Carbon & Random & $2.10\%$ & $[1.75\%,\,2.42\%]$ & $2.42\%$ \\
Carbon & FPS/DPP & $1.94\%$ & $[1.55\%,\,2.17\%]$ & $2.17\%$ \\
Carbon & QUESTS-MSC & $5.05\%$ & $[4.80\%,\,5.76\%]$ & $5.94\%$ \\
Carbon & Response witness & $57.85\%$ & $[52.04\%,\,60.73\%]$ & $60.73\%$ \\
Carbon & Response repair & $1.92\%$ & $[1.41\%,\,2.35\%]$ & $2.35\%$ \\
rMD17 full & Random & $1.76\%$ & $[1.25\%,\,2.10\%]$ & $1.76\%$ \\
rMD17 full & FPS/DPP & $1.56\%$ & $[1.10\%,\,1.90\%]$ & $1.56\%$ \\
rMD17 full & QUESTS-MSC & $6.94\%$ & $[6.47\%,\,7.33\%]$ & $6.94\%$ \\
rMD17 full & Response witness & $24.20\%$ & $[22.92\%,\,24.29\%]$ & $22.72\%$ \\
rMD17 full & Response repair & $1.56\%$ & $[1.10\%,\,1.90\%]$ & $1.56\%$ \\
\bottomrule
\end{tabular}%
}
\vspace{1pt}
\noindent\textit{Panel C: five-seed high-force-tail gaps.}\\[2pt]
\resizebox{\textwidth}{!}{%
\begin{tabular}{llllllll}
\toprule
Dataset & Selector & Seed 1 & Seed 2 & Seed 3 & Seed 4 & Seed 5 & Mean \\
\midrule
Carbon & Random & $10.08\%$ & $10.69\%$ & $12.26\%$ & $11.18\%$ & $13.04\%$ & $11.45\%$ \\
Carbon & FPS/DPP & $10.35\%$ & $9.61\%$ & $10.98\%$ & $9.85\%$ & $11.81\%$ & $10.52\%$ \\
Carbon & QUESTS-MSC & $17.54\%$ & $19.15\%$ & $16.61\%$ & $16.76\%$ & $17.63\%$ & $17.54\%$ \\
Carbon & Response witness & $56.24\%$ & $56.61\%$ & $54.19\%$ & $52.23\%$ & $56.98\%$ & $55.25\%$ \\
Carbon & Response repair & $10.38\%$ & $9.55\%$ & $11.23\%$ & $10.03\%$ & $11.82\%$ & $10.60\%$ \\
rMD17 & Random & $9.68\%$ & $9.57\%$ & $9.99\%$ & $10.14\%$ & $10.03\%$ & $9.88\%$ \\
rMD17 & FPS/DPP & $9.91\%$ & $8.70\%$ & $8.97\%$ & $9.02\%$ & $8.96\%$ & $9.11\%$ \\
rMD17 & QUESTS-MSC & $25.52\%$ & $22.69\%$ & $25.52\%$ & $22.57\%$ & $25.29\%$ & $24.32\%$ \\
rMD17 & Response witness & $21.34\%$ & $24.67\%$ & $23.46\%$ & $23.61\%$ & $24.50\%$ & $23.52\%$ \\
rMD17 & Response repair & $9.91\%$ & $8.70\%$ & $8.97\%$ & $9.02\%$ & $8.96\%$ & $9.11\%$ \\
\bottomrule
\end{tabular}%
}
\vspace{1pt}
\noindent\textit{Panel D: five-seed High-novelty mask gaps.}\\[2pt]
\resizebox{\textwidth}{!}{%
\begin{tabular}{llllllll}
\toprule
Dataset & Selector & Seed 1 & Seed 2 & Seed 3 & Seed 4 & Seed 5 & Mean \\
\midrule
Carbon & Random & $5.19\%$ & $5.62\%$ & $6.28\%$ & $6.11\%$ & $6.90\%$ & $6.02\%$ \\
Carbon & FPS/DPP & $5.29\%$ & $5.08\%$ & $5.66\%$ & $5.44\%$ & $6.28\%$ & $5.55\%$ \\
Carbon & QUESTS-MSC & $11.11\%$ & $12.19\%$ & $10.32\%$ & $10.55\%$ & $11.25\%$ & $11.08\%$ \\
Carbon & Response witness & $54.96\%$ & $55.40\%$ & $53.31\%$ & $51.54\%$ & $55.81\%$ & $54.20\%$ \\
Carbon & Response repair & $5.61\%$ & $5.26\%$ & $6.19\%$ & $5.85\%$ & $6.59\%$ & $5.90\%$ \\
rMD17 & Random & $4.16\%$ & $4.32\%$ & $4.45\%$ & $4.51\%$ & $4.36\%$ & $4.36\%$ \\
rMD17 & FPS/DPP & $4.23\%$ & $3.92\%$ & $3.98\%$ & $4.01\%$ & $3.91\%$ & $4.01\%$ \\
rMD17 & QUESTS-MSC & $11.34\%$ & $10.69\%$ & $11.94\%$ & $10.04\%$ & $11.62\%$ & $11.13\%$ \\
rMD17 & Response witness & $20.15\%$ & $22.37\%$ & $21.59\%$ & $21.65\%$ & $22.55\%$ & $21.66\%$ \\
rMD17 & Response repair & $4.23\%$ & $3.92\%$ & $3.98\%$ & $4.01\%$ & $3.91\%$ & $4.01\%$ \\
\bottomrule
\end{tabular}%
}
\end{table*}

FPS/DPP has smaller 5\% gaps than Random for all four endpoints in both datasets (Supplementary Table~\ref{tab:si_outcome_gap}). Fully enumerated paired-seed bootstrap intervals for Random minus FPS/DPP are above zero in all eight comparisons (main-text Table~\ref{M-tab:paired_coverage}). These intervals describe variation over five training seeds. QUESTS-MSC and response-witness selection have larger gaps. Carbon repair remains close to FPS/DPP, with all four direct-effect intervals spanning zero; rMD17 repair reproduces FPS/DPP because no replacement occurs at this budget.

The budget comparison extends the baseline, response-witness and repair results from the primary 5\% setting to 1\% and 20\%. These additional points test whether the selector ordering persists as the retained subset grows.

\begin{figure*}[!tp]
\centering
\includegraphics[width=172mm]{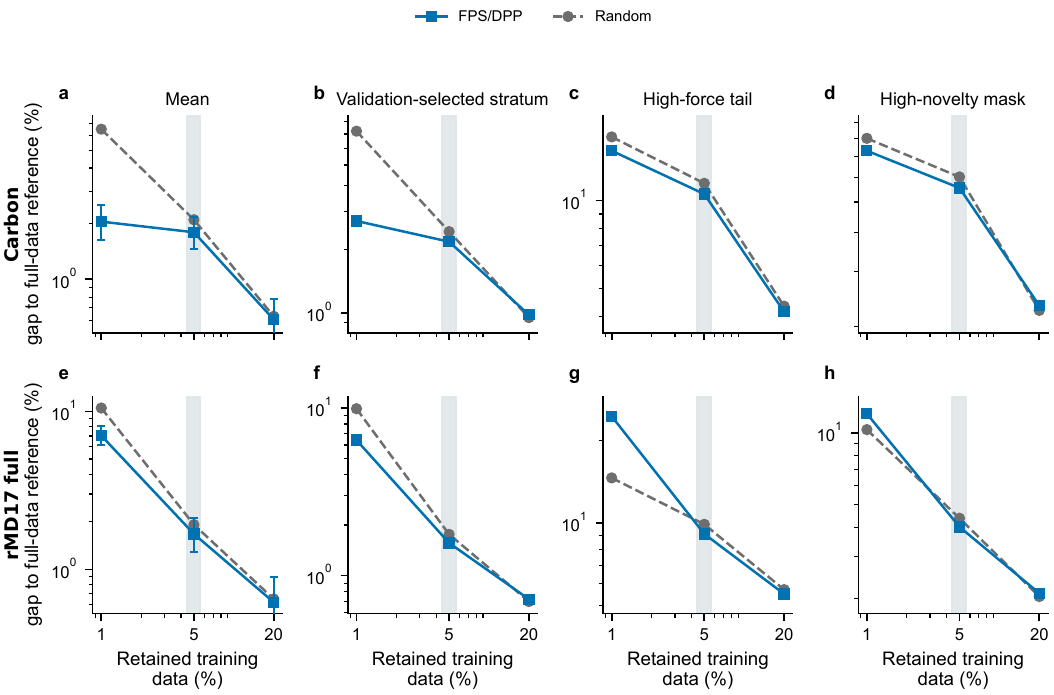}
\caption{\textbf{Random and FPS/DPP outcome gaps across budget.} Carbon and rMD17 panels compare the two baselines at 1\%, 5\% and 20\% for four absolute gap endpoints. Points are five-seed means and lines are visual guides; the pale band marks the primary 5\% budget, not an uncertainty interval. Error bars in the FPS/DPP mean panels show the reported 95\% intervals.}
\label{fig:si_budget_sensitivity}
\end{figure*}

Selector ordering changes with budget. FPS/DPP has lower 5\% gaps than Random on all four endpoints in both datasets, but the 1\% rMD17 comparison is mixed: FPS/DPP is closer on the Mean and Validation-selected stratum endpoints, whereas Random is closer on the High-force tail and High-novelty mask endpoints. At 20\%, the two baselines are close. Response-witness gaps are largest at 1\% and fall below FPS/DPP at 20\% across all four endpoints in both datasets. Fully enumerated paired-seed intervals support the 20\% reversal; the 1\% comparison describes the ordering of the five-seed means.

\begin{table*}[!tp]
\centering
\scriptsize
\caption{\textbf{Direct outcome gaps across budget.} Values are five-seed means relative to the full-data training outcome; lower absolute gaps indicate closer agreement. ``Validation-selected stratum'' denotes the coded stratum fixed on validation data.}
\label{tab:si_budget_sensitivity}
\resizebox{\textwidth}{!}{%
\begin{tabular}{lllllll}
\toprule
Dataset & Budget & Selector & Mean & Validation-selected stratum & High-force tail & High-novelty mask \\
\midrule
Carbon & $1\%$ & Random & $6.50\%$ & $7.22\%$ & $16.50\%$ & $8.00\%$ \\
Carbon & $5\%$ & Random & $2.10\%$ & $2.42\%$ & $11.45\%$ & $6.02\%$ \\
Carbon & $20\%$ & Random & $0.63\%$ & $0.95\%$ & $4.35\%$ & $2.25\%$ \\
Carbon & $1\%$ & FPS/DPP & $2.05\%$ & $2.72\%$ & $14.80\%$ & $7.30\%$ \\
Carbon & $5\%$ & FPS/DPP & $1.80\%$ & $2.17\%$ & $10.52\%$ & $5.55\%$ \\
Carbon & $20\%$ & FPS/DPP & $0.60\%$ & $0.98\%$ & $4.16\%$ & $2.33\%$ \\
Carbon & $1\%$ & QUESTS-MSC & $42.98\%$ & $44.76\%$ & $65.59\%$ & $45.99\%$ \\
Carbon & $5\%$ & QUESTS-MSC & $4.99\%$ & $5.94\%$ & $17.54\%$ & $11.08\%$ \\
Carbon & $20\%$ & QUESTS-MSC & $1.01\%$ & $1.41\%$ & $6.19\%$ & $4.01\%$ \\
Carbon & $1\%$ & Response witness & $212.31\%$ & $213.22\%$ & $222.86\%$ & $189.40\%$ \\
Carbon & $5\%$ & Response witness & $57.27\%$ & $60.73\%$ & $55.25\%$ & $54.20\%$ \\
Carbon & $20\%$ & Response witness & $0.14\%$ & $0.04\%$ & $0.93\%$ & $0.27\%$ \\
Carbon & $1\%$ & Response repair & $2.57\%$ & $2.92\%$ & $7.66\%$ & $3.90\%$ \\
Carbon & $5\%$ & Response repair & $1.86\%$ & $2.35\%$ & $10.60\%$ & $5.90\%$ \\
Carbon & $20\%$ & Response repair & $0.60\%$ & $0.98\%$ & $4.16\%$ & $2.33\%$ \\
rMD17 full & $1\%$ & Random & $10.51\%$ & $9.87\%$ & $14.61\%$ & $10.31\%$ \\
rMD17 full & $5\%$ & Random & $1.92\%$ & $1.76\%$ & $9.88\%$ & $4.36\%$ \\
rMD17 full & $20\%$ & Random & $0.65\%$ & $0.70\%$ & $5.70\%$ & $2.04\%$ \\
rMD17 full & $1\%$ & FPS/DPP & $7.03\%$ & $6.41\%$ & $24.55\%$ & $12.03\%$ \\
rMD17 full & $5\%$ & FPS/DPP & $1.68\%$ & $1.56\%$ & $9.11\%$ & $4.01\%$ \\
rMD17 full & $20\%$ & FPS/DPP & $0.62\%$ & $0.72\%$ & $5.50\%$ & $2.11\%$ \\
rMD17 full & $1\%$ & QUESTS-MSC & $95.16\%$ & $95.92\%$ & $134.58\%$ & $98.03\%$ \\
rMD17 full & $5\%$ & QUESTS-MSC & $6.78\%$ & $6.94\%$ & $24.32\%$ & $11.13\%$ \\
rMD17 full & $20\%$ & QUESTS-MSC & $1.80\%$ & $2.14\%$ & $11.10\%$ & $4.32\%$ \\
rMD17 full & $1\%$ & Response witness & $190.79\%$ & $184.26\%$ & $181.53\%$ & $184.24\%$ \\
rMD17 full & $5\%$ & Response witness & $24.13\%$ & $22.72\%$ & $23.52\%$ & $21.66\%$ \\
rMD17 full & $20\%$ & Response witness & $0.11\%$ & $0.16\%$ & $0.72\%$ & $0.13\%$ \\
rMD17 full & $1\%$ & Response repair & $6.25\%$ & $5.58\%$ & $21.89\%$ & $10.86\%$ \\
rMD17 full & $5\%$ & Response repair & $1.68\%$ & $1.56\%$ & $9.11\%$ & $4.01\%$ \\
rMD17 full & $20\%$ & Response repair & $0.62\%$ & $0.72\%$ & $5.50\%$ & $2.11\%$ \\
\bottomrule
\end{tabular}%
}
\end{table*}

The 1--20\% comparison places the primary 5\% result on a budget curve.

Tables~\ref{tab:si_outcome_gap} and~\ref{tab:si_budget_sensitivity} connect training-seed variation with the cross-budget reversal. At 20\%, response witness has both smaller absolute gaps and lower direct DFT risk than FPS/DPP across both datasets and all four endpoints. Supplementary Table~\ref{tab:si_20pct_paired} gives the five paired seeds, fully enumerated paired-bootstrap intervals and paired-$t$ sensitivity intervals.

\subsection{Confidence-interval precision and ranking stability}
\label{sec:si_detectability_stability}

The 95\% confidence-interval half-width, $(\mathrm{CI}_{\mathrm{high}}-\mathrm{CI}_{\mathrm{low}})/2$, summarises the precision of each reported contrast. Panels A and C of Supplementary Table~\ref{tab:si_precision_summary} use paired training-seed intervals; Panel B retains the grouped response-effect intervals.

\begin{table*}[!tp]
\centering
\scriptsize
\caption{\textbf{Confidence intervals for outcome contrasts at the 5\% budget.} Panel A reports response-witness minus FPS/DPP absolute-gap differences; Panel C reports Random minus FPS/DPP absolute-gap differences. Both enumerate all $5^5$ paired training-seed bootstrap resamples. Panel B reports relative response-risk effects with the grouped intervals from Supplementary Table~\ref{tab:si_response_repair_summary}. Panels A and C are in percentage points; Panel B is in percent. The CI half-width is one half of the unrounded 95\% interval width.}
\label{tab:si_precision_summary}
\resizebox{\textwidth}{!}{%
\begin{tabular}{lllllll}
\toprule
Panel & Dataset/method & Endpoint & Change vs FPS/DPP & 95\% CI & CI half-width & Interpretation \\
\midrule
A & Carbon & Mean gap & $+55.474$ & $[+53.908,\,+56.795]$ & $1.443$ & Witness farther from full-data outcome \\
A & Carbon & Validation-selected stratum gap & $+58.565$ & $[+56.943,\,+59.875]$ & $1.466$ & Witness farther from full-data outcome \\
A & Carbon & High-force tail gap & $+44.728$ & $[+43.245,\,+46.189]$ & $1.472$ & Witness farther from full-data outcome \\
A & Carbon & High-novelty mask gap & $+48.654$ & $[+47.124,\,+49.902]$ & $1.389$ & Witness farther from full-data outcome \\
A & rMD17 & Mean gap & $+22.458$ & $[+21.492,\,+23.219]$ & $0.863$ & Witness farther from full-data outcome \\
A & rMD17 & Validation-selected stratum gap & $+21.156$ & $[+20.172,\,+21.888]$ & $0.858$ & Witness farther from full-data outcome \\
A & rMD17 & High-force tail gap & $+14.404$ & $[+12.863,\,+15.591]$ & $1.364$ & Witness farther from full-data outcome \\
A & rMD17 & High-novelty mask gap & $+17.652$ & $[+16.764,\,+18.395]$ & $0.816$ & Witness farther from full-data outcome \\
B & Carbon witness & Metadata-score mask & $+59.446\%$ & $[+49.127\%,\,+71.695\%]$ & $11.284\%$ & Higher risk \\
B & Carbon witness & High-force tail & $+40.471\%$ & $[+33.509\%,\,+48.234\%]$ & $7.363\%$ & Higher risk \\
B & Carbon witness & High-response cell & $+6.329\%$ & $[+2.185\%,\,+13.521\%]$ & $5.668\%$ & Higher risk \\
B & Carbon witness & High-novelty mask & $+46.095\%$ & $[+38.435\%,\,+54.332\%]$ & $7.948\%$ & Higher risk \\
B & Carbon repair & Metadata-score mask & $-0.088\%$ & $[-0.365\%,\,+0.178\%]$ & $0.272\%$ & CI includes zero \\
B & Carbon repair & High-force tail & $+0.076\%$ & $[-0.395\%,\,+0.570\%]$ & $0.483\%$ & CI includes zero \\
B & Carbon repair & High-response cell & $-1.446\%$ & $[-3.484\%,\,+0.672\%]$ & $2.078\%$ & CI includes zero \\
B & Carbon repair & High-novelty mask & $+0.333\%$ & $[-0.049\%,\,+0.724\%]$ & $0.386\%$ & CI includes zero \\
B & rMD17 witness & Metadata-score mask & $+58.515\%$ & $[+50.864\%,\,+67.402\%]$ & $8.269\%$ & Higher risk \\
B & rMD17 witness & High-force tail & $+13.200\%$ & $[+5.481\%,\,+21.187\%]$ & $7.853\%$ & Higher risk \\
B & rMD17 witness & High-response cell & $-7.872\%$ & $[-16.547\%,\,-1.607\%]$ & $7.470\%$ & Lower risk \\
B & rMD17 witness & High-novelty mask & $+16.971\%$ & $[+12.175\%,\,+21.919\%]$ & $4.872\%$ & Higher risk \\
B & rMD17 repair & Metadata-score mask & $0.000\%$ & $[0.000\%,\,0.000\%]$ & $0.000\%$ & Identical subset \\
B & rMD17 repair & High-force tail & $0.000\%$ & $[0.000\%,\,0.000\%]$ & $0.000\%$ & Identical subset \\
B & rMD17 repair & High-response cell & $0.000\%$ & $[0.000\%,\,0.000\%]$ & $0.000\%$ & Identical subset \\
B & rMD17 repair & High-novelty mask & $0.000\%$ & $[0.000\%,\,0.000\%]$ & $0.000\%$ & Identical subset \\
\midrule
C & Carbon & Mean gap & $+0.303$ & $[+0.111,\,+0.411]$ & $0.150$ & FPS/DPP closer \\
C & Carbon & Validation-selected stratum gap & $+0.251$ & $[+0.125,\,+0.329]$ & $0.102$ & FPS/DPP closer \\
C & Carbon & High-force tail gap & $+0.931$ & $[+0.321,\,+1.291]$ & $0.485$ & FPS/DPP closer \\
C & Carbon & High-novelty mask gap & $+0.470$ & $[+0.182,\,+0.640]$ & $0.229$ & FPS/DPP closer \\
C & rMD17 & Mean gap & $+0.244$ & $[+0.084,\,+0.356]$ & $0.136$ & FPS/DPP closer \\
C & rMD17 & Validation-selected stratum gap & $+0.197$ & $[+0.085,\,+0.263]$ & $0.089$ & FPS/DPP closer \\
C & rMD17 & High-force tail gap & $+0.769$ & $[+0.259,\,+1.079]$ & $0.410$ & FPS/DPP closer \\
C & rMD17 & High-novelty mask gap & $+0.350$ & $[+0.138,\,+0.478]$ & $0.170$ & FPS/DPP closer \\
\bottomrule
\end{tabular}%
}
\end{table*}

Table~\ref{tab:si_precision_summary} reports the recalculated CI half-widths. Carbon witness intervals indicate higher risk at all four endpoints. The rMD17 witness has one lower-risk and three higher-risk intervals, Carbon repair intervals all include zero, and rMD17 repair is identical to FPS/DPP at 5\%. The paired-seed Random-minus-FPS/DPP intervals are above zero for all eight absolute-gap endpoints; they quantify variation across five training seeds rather than observation-level groups.

The top-budget margin describes score separation at the coverage-selection boundary. For a standardised raw coverage score $s_{\mathrm{cov}}(x)$, transformed with parameters fitted on the training pool, the empirical margin is
\[
\widehat{\gamma}_B=s_{\mathrm{cov}}\!\left(x_{(B)}\right)-s_{\mathrm{cov}}\!\left(x_{(B+1)}\right),
\]
where $x_{(B)}$ and $x_{(B+1)}$ are the structures at the budget boundary under the coverage ranking. In an ideal calibrated-score case, a perturbation smaller than $\widehat{\gamma}_B/2$ cannot change the top-$B$ set except for ties. Here the margin is only a descriptive boundary diagnostic and is interpreted alongside rank correlation, Jaccard overlap and group-discrimination performance.

\begin{table*}[!tp]
\centering
\small
\caption{\textbf{Top-budget stability and coverage alignment.}
Panel A reports top-budget margins for the standardised Gaussian QUESTS coverage score and Jaccard overlap with the learned-metric selection. Panel B reports the 5\% comparison. Margins are descriptive boundary diagnostics rather than formal guarantees.}
\label{tab:si_gamma_stability}
\adjustbox{max width=\textwidth}{%
\begin{tabular}{l c c c l}
\toprule
\multicolumn{5}{l}{\emph{Panel A: boundary-margin estimates for Gaussian QUESTS}} \\
\midrule
Dataset & Budget & $\widehat{\gamma}_B$ & Jaccard vs learned metric & Stability interpretation \\
\midrule
Carbon & $1\%$ & $2.022\times10^{-3}$ & $0.667$ & Frozen boundary diagnostic \\
Carbon & $2\%$ & $6.238\times10^{-4}$ & $0.667$ & Frozen boundary diagnostic \\
Carbon & $5\%$ & $5.501\times10^{-4}$ & $0.754$ & Primary-budget diagnostic \\
Carbon & $10\%$ & $1.060\times10^{-3}$ & $0.835$ & Frozen boundary diagnostic \\
Carbon & $20\%$ & $3.719\times10^{-5}$ & $0.852$ & Frozen boundary diagnostic \\
rMD17 full & $1\%$ & $7.458\times10^{-4}$ & $0.538$ & Frozen boundary diagnostic \\
rMD17 full & $2\%$ & $3.078\times10^{-3}$ & $0.667$ & Frozen boundary diagnostic \\
rMD17 full & $5\%$ & $8.435\times10^{-4}$ & $0.724$ & Primary-budget diagnostic \\
rMD17 full & $10\%$ & $1.924\times10^{-3}$ & $0.786$ & Frozen boundary diagnostic \\
rMD17 full & $20\%$ & $1.687\times10^{-4}$ & $0.878$ & Frozen boundary diagnostic \\
HEA & $1\%$ & $1.144\times10^{-3}$ & $0.538$ & Frozen boundary diagnostic \\
HEA & $2\%$ & $3.291\times10^{-3}$ & $0.600$ & Frozen boundary diagnostic \\
HEA & $5\%$ & $7.504\times10^{-4}$ & $0.667$ & Primary-budget diagnostic \\
HEA & $10\%$ & $3.102\times10^{-3}$ & $0.739$ & Frozen boundary diagnostic \\
HEA & $20\%$ & $1.546\times10^{-4}$ & $0.794$ & Frozen boundary diagnostic \\
\midrule
\multicolumn{5}{l}{\emph{Panel B: primary-budget learned-metric alignment}} \\
\midrule
Dataset & Spearman $\rho_S$ & Top $5\%$ Jaccard & Learned$-$Gaussian $\Delta$AUROC & Interpretation \\
\midrule
Carbon & $0.990$ & $0.754$ & $+0.002$ [$-0.003$, $+0.008$] & Coverage-aligned; no resolved AUROC gain \\
rMD17 full & $0.990$ & $0.724$ & $-0.002$ [$-0.007$, $+0.003$] & Coverage-aligned; no resolved AUROC gain \\
HEA & $0.980$ & $0.667$ & $-0.002$ [$-0.013$, $+0.007$] & Coverage-aligned; no resolved AUROC gain \\
\bottomrule
\end{tabular}%
}
\end{table*}

The learned metric changes membership near the budget boundary, particularly for HEA at small budgets (Supplementary Table~\ref{tab:si_gamma_stability}). Full-ranking correlations nevertheless remain close to unity, and leave-one-group-out AUROC differences remain centred near zero. The boundary margins complement these observed rank and overlap statistics.

\subsection{Learned-metric positive-control capacity test}
\label{sec:si_learned_metric_positive_control}

The strong rank agreement on the physical datasets motivates a capacity test of the bounded metric. A constructed positive control supplies an independent value channel by injecting a known signal into generated feature vectors. This tests whether the same parameterisation can depart from Gaussian QUESTS coverage.

We drew 2,000 eight-dimensional Gaussian geometry vectors and assigned an independent high-value label with $15\%$ prevalence. A stratified frozen split placed 1,000 samples in a metric-fit pool and 1,000 in an evaluation pool. Coverage novelty $c(x)$ was the mean five-nearest-neighbour distance from each evaluation sample to the fit pool. The structured channel was $v(x)=2y(x)+\epsilon$, with $\epsilon\sim\mathcal{N}(0,1)$; the null channel contained no label signal, and the shuffled control randomly permuted the structured channel. Pair targets and metric weights were fitted only on the fit pool. All ranks, top-budget overlaps and AUROCs were computed on the evaluation pool against the frozen fit reference. Intervals are percentile intervals from 2,000 paired bootstrap resamples of the evaluation structures.

\begin{figure*}[!tp]
\centering
\includegraphics[width=172mm]{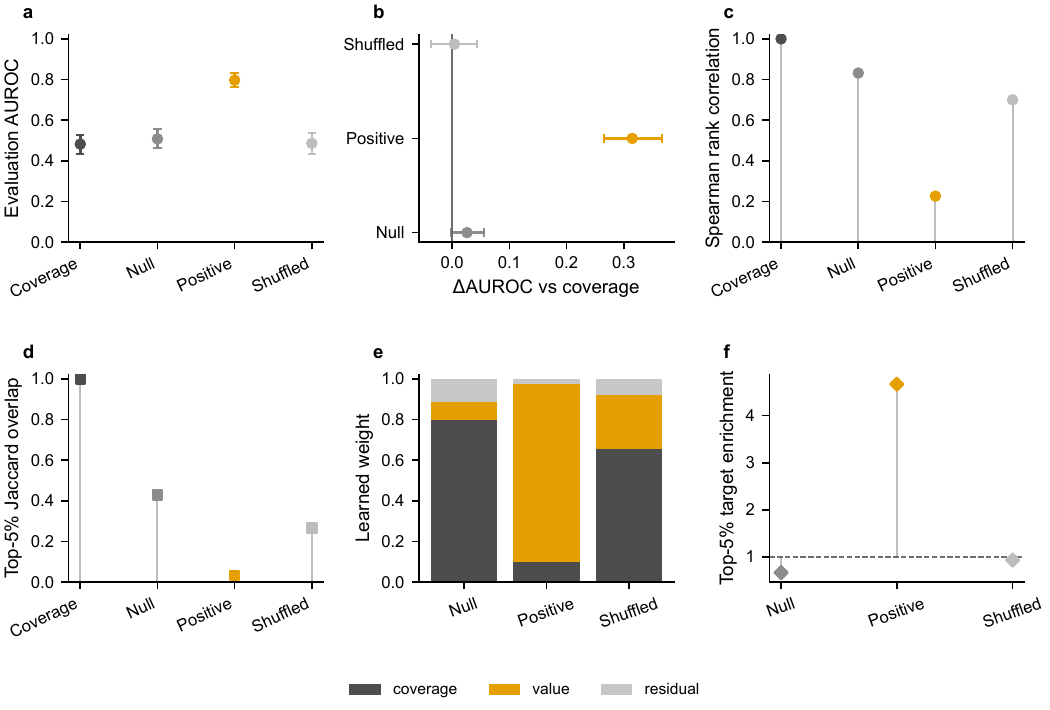}
\caption{\textbf{The learned metric responds to an injected independent value channel.} \textbf{a}, Evaluation AUROC for the coverage baseline and learned metric under null, positive and shuffled controls. \textbf{b}, Learned-minus-coverage $\Delta$AUROC with paired-bootstrap 95\% confidence intervals. \textbf{c}, Spearman rank correlation with the coverage ordering. \textbf{d}, Top-5\% Jaccard overlap with the coverage subset. \textbf{e}, Learned channel weights under the three control settings. \textbf{f}, Top-5\% enrichment of the target group; the dashed line marks no enrichment. This methodological control uses generated features with a known injected signal and disjoint fitting and evaluation pools.}
\label{fig:si_learned_metric_positive_control}
\end{figure*}

The injected channel changes the learned ranking (Supplementary Fig.~\ref{fig:si_learned_metric_positive_control}). Under the null control, the metric remains coverage-dominated, with Spearman $\rho_{S}=0.832$, top-5\% Jaccard overlap of 0.429 and coverage-channel weight 0.797. The positive control shifts weight to the injected value channel, which receives 0.878; $\rho_S$ falls to 0.226 and Jaccard overlap to 0.031. Evaluation AUROC increases from 0.482 for coverage to 0.797 for the learned metric. Shuffling the channel gives AUROC 0.486 and a paired $\Delta$AUROC interval that includes zero.

\begin{table*}[!tp]
\centering
\small
\caption{\textbf{Learned-metric positive-control statistics.}
Null, positive and shuffled channels test the metric's response to a known signal independent of geometry. Metric fitting and evaluation use disjoint frozen pools. Spearman $\rho_S$ and top-5\% Jaccard are computed relative to nearest-neighbour coverage; $\Delta$AUROC is paired against the coverage baseline.}
\label{tab:si_learned_metric_positive_control}

\noindent\textit{Panel A: rank agreement and injected-group detection.}\\[2pt]
{\setlength{\tabcolsep}{4pt}%
\resizebox{\textwidth}{!}{%
\begin{tabular}{p{2.8cm}p{3.8cm}cccp{2.7cm}p{3.4cm}}
\toprule
Setting
& Value-channel condition
& $\rho(c,v)$
& Spearman $\rho_S$
& Top $5\%$ Jaccard
& AUROC
& $\Delta$AUROC \\
\midrule
Coverage baseline
& Coverage baseline
& $-0.0003$
& $1.000$
& $1.00$
& $0.482$ [$0.433$, $0.529$]
& $0.000$
\\
Learned metric, null
& No structured independent value channel
& $-0.0003$
& $0.832$
& $0.429$
& $0.508$ [$0.461$, $0.555$]
& $+0.026$ [$-0.003$, $+0.056$]
\\
Learned metric, positive control
& Structured injected value channel
& $-0.0003$
& $0.226$
& $0.031$
& $0.797$ [$0.761$, $0.831$]
& $+0.315$ [$+0.265$, $+0.367$]
\\
Learned metric, shuffled control
& Shuffled injected channel
& $-0.0003$
& $0.700$
& $0.266$
& $0.486$ [$0.432$, $0.537$]
& $+0.004$ [$-0.038$, $+0.043$]
\\
\bottomrule
\end{tabular}%
}
}

\vspace{0.8em}

\noindent\textit{Panel B: learned channel weights and interpretations.}\\[2pt]
\begin{tabular}{lcccp{6.7cm}}
\toprule
Setting
& $w_{\mathrm{cov}}$
& $w_{\mathrm{val}}$
& $w_{\mathrm{res}}$
& Interpretation \\
\midrule
Coverage baseline
& $1.000$
& $0.000$
& $0.000$
& Frozen reference comparator \\
Learned metric, null
& $0.797$
& $0.091$
& $0.111$
& No statistically resolved AUROC gain \\
Learned metric, positive control
& $0.098$
& $0.878$
& $0.024$
& Can reorder and enrich target group \\
Learned metric, shuffled control
& $0.656$
& $0.263$
& $0.081$
& No reliable AUROC gain \\
\bottomrule
\end{tabular}
\end{table*}

The metric detects the structured channel in this constructed setting (Supplementary Table~\ref{tab:si_learned_metric_positive_control}). The paired $\Delta$AUROC intervals include zero for the null and shuffled controls. With the injected signal, the value channel receives weight 0.878 and the selected subset enriches the high-value group by $4.67\times$. This control shows that the bounded metric can reorder selections in response to the specified independent signal; the physical datasets assess its behaviour with measured energy, force and structural attributes.

\subsection{Uncertainty available for each outcome contrast}
\label{sec:si_interval_availability}

Supplementary Table~\ref{tab:si_interval_map} maps each outcome comparison to its uncertainty estimate. Grouped intervals describe variation across scientific groups, whereas paired-seed intervals describe variation across matched training seeds. Comparisons without direct contrast intervals are described by five-seed means.

\begin{table*}[!tp]
\centering
\footnotesize
\setlength{\tabcolsep}{4pt}
\caption{\textbf{Uncertainty estimates for the outcome comparisons.} Every budget comprises five training seeds. Grouped intervals resample the stated scientific groups with paired seeds; paired-seed intervals resample the five matched training seeds. The primary 5\% coverage contrast and the 20\% response crossover have paired-seed intervals. The 1\% comparison reports the ordering of five-seed means.}
\label{tab:si_interval_map}
\begin{tabular}{P{3.7cm}P{5.0cm}P{4.2cm}P{4.55cm}}
\toprule
Contrast & Evidence at 5\% & Evidence at 1\% and 20\% & Supported interpretation \\
\midrule
Random $-$ FPS/DPP absolute gap & Five paired seeds for all four endpoints; fully enumerated $5^5$ paired-seed bootstrap & Five-seed means are reported at both budgets; no additional Random--FPS/DPP interval is shown & Direct paired interval at 5\%; descriptive ordering at 1\% and 20\% \\
Witness $-$ FPS/DPP absolute gap & Five paired seeds for all four endpoints; $5^5$ paired-seed bootstrap & Five-seed means at 1\%; paired-seed interval at 20\% & Paired-seed contrasts at 5\% and 20\% \\
Witness or repair $-$ FPS/DPP risk & Grouped effect intervals for all four endpoints & Five-seed means are reported at both budgets; witness direct DFT-risk intervals are shown at 20\% & Endpoint trade-offs at 5\%; lower witness DFT risk across all eight 20\% comparisons \\
Witness signed gap from full-data risk & Not used for the primary 5\% contrast & Five matched seeds with fully enumerated paired bootstrap and paired-$t$ intervals at 20\% & Witness risk below full-data risk in six of eight 20\% comparisons \\
FPS/DPP gap to full-data outcome & Reported mean-gap intervals & Reported mean-gap intervals at 1\% and 20\% & Uncertainty in gap magnitude; direct 20\% direction is assessed through paired DFT risks \\
Repair $-$ FPS/DPP absolute gap & Five paired-seed summaries; grouped risk effects & Five-seed means are reported at both budgets; no additional direct interval is shown & Selection identity is established only for rMD17 at 5\% \\
QUESTS-MSC direct contrast & Five-seed summaries; no direct contrast interval reported & Five-seed means are reported at both budgets; no direct contrast interval is shown & Descriptive secondary coverage anchor \\
\bottomrule
\end{tabular}
\end{table*}

\begin{table*}[!tp]
\centering
\scriptsize
\caption{\textbf{Training-seed sensitivity for the primary 5\% coverage comparison and the 20\% response crossover.} Panel A compares Random with FPS/DPP using absolute-gap differences in percentage points. Panel B lists response-witness minus FPS/DPP held-out DFT-risk differences in meV~$\text{\AA}^{-1}$ for each matched seed. Panel C summarises the corresponding witness--FPS/DPP absolute-gap contrast plotted in Fig.~2c. Panel D reports the signed response-witness gap from the matched full-data DFT risk. Bootstrap intervals enumerate all $5^5$ paired-seed resamples; paired-$t$ intervals use four degrees of freedom. Positive values in Panel A favour FPS/DPP. Negative values in Panels B and C favour response witness; negative values in Panel D indicate lower witness risk than the full-data reference.}
\label{tab:si_20pct_paired}
\noindent\textit{Panel A: Random minus FPS/DPP absolute-gap differences at 5\%.}\\[2pt]
\begin{tabular}{llrrr}
\toprule
Dataset & Endpoint & Mean difference & Bootstrap 95\% interval & Paired-$t$ 95\% interval \\
\midrule
Carbon & Mean & $+0.303$ & $[+0.111,+0.411]$ & $[+0.038,+0.569]$ \\
Carbon & Validation-selected stratum & $+0.251$ & $[+0.125,+0.329]$ & $[+0.075,+0.427]$ \\
Carbon & High-force tail & $+0.931$ & $[+0.321,+1.291]$ & $[+0.090,+1.772]$ \\
Carbon & High-novelty mask & $+0.470$ & $[+0.182,+0.640]$ & $[+0.070,+0.870]$ \\
rMD17 full & Mean & $+0.244$ & $[+0.084,+0.356]$ & $[+0.023,+0.466]$ \\
rMD17 full & Validation-selected stratum & $+0.197$ & $[+0.085,+0.263]$ & $[+0.042,+0.351]$ \\
rMD17 full & High-force tail & $+0.769$ & $[+0.259,+1.079]$ & $[+0.065,+1.473]$ \\
rMD17 full & High-novelty mask & $+0.350$ & $[+0.138,+0.478]$ & $[+0.055,+0.645]$ \\
\bottomrule
\end{tabular}

\vspace{0.7em}
\noindent\textit{Panel B: response witness minus FPS/DPP direct DFT-risk differences at 20\%.}\\[2pt]
\resizebox{\textwidth}{!}{%
\begin{tabular}{llrrrrrrr}
\toprule
Dataset & Endpoint & Seed 1 & Seed 2 & Seed 3 & Seed 4 & Seed 5 & Bootstrap 95\% interval & Paired-$t$ 95\% interval \\
\midrule
Carbon & Mean & $-0.147$ & $-0.177$ & $-0.201$ & $-0.136$ & $-0.173$ & $[-0.186,-0.146]$ & $[-0.199,-0.134]$ \\
Carbon & Validation-selected stratum & $-0.300$ & $-0.350$ & $-0.344$ & $-0.287$ & $-0.345$ & $[-0.347,-0.301]$ & $[-0.362,-0.288]$ \\
Carbon & High-force tail & $-1.867$ & $-1.914$ & $-2.207$ & $-1.712$ & $-1.717$ & $[-2.050,-1.745]$ & $[-2.134,-1.632]$ \\
Carbon & High-novelty mask & $-1.033$ & $-1.091$ & $-1.124$ & $-0.970$ & $-1.061$ & $[-1.099,-1.007]$ & $[-1.129,-0.983]$ \\
rMD17 full & Mean & $-0.180$ & $-0.152$ & $-0.195$ & $-0.156$ & $-0.134$ & $[-0.183,-0.146]$ & $[-0.193,-0.134]$ \\
rMD17 full & Validation-selected stratum & $-0.222$ & $-0.186$ & $-0.224$ & $-0.192$ & $-0.183$ & $[-0.217,-0.186]$ & $[-0.226,-0.177]$ \\
rMD17 full & High-force tail & $-1.546$ & $-1.465$ & $-1.514$ & $-1.434$ & $-1.484$ & $[-1.524,-1.456]$ & $[-1.543,-1.434]$ \\
rMD17 full & High-novelty mask & $-0.546$ & $-0.533$ & $-0.573$ & $-0.529$ & $-0.520$ & $[-0.557,-0.526]$ & $[-0.566,-0.515]$ \\
\bottomrule
\end{tabular}%
}

\vspace{0.7em}
\noindent\textit{Panel C: response witness minus FPS/DPP absolute-gap differences at 20\% (percentage points).}\\[2pt]
\begin{tabular}{llrrr}
\toprule
Dataset & Endpoint & Mean difference & Bootstrap 95\% interval & Paired-$t$ 95\% interval \\
\midrule
Carbon & Mean & $-0.461$ & $[-0.514,-0.404]$ & $[-0.550,-0.371]$ \\
Carbon & Validation-selected stratum & $-0.940$ & $[-1.003,-0.870]$ & $[-1.046,-0.833]$ \\
Carbon & High-force tail & $-3.234$ & $[-3.341,-3.135]$ & $[-3.398,-3.070]$ \\
Carbon & High-novelty mask & $-2.058$ & $[-2.096,-1.996]$ & $[-2.142,-1.974]$ \\
rMD17 full & Mean & $-0.513$ & $[-0.554,-0.485]$ & $[-0.569,-0.457]$ \\
rMD17 full & Validation-selected stratum & $-0.563$ & $[-0.596,-0.536]$ & $[-0.611,-0.514]$ \\
rMD17 full & High-force tail & $-4.780$ & $[-4.936,-4.653]$ & $[-4.994,-4.567]$ \\
rMD17 full & High-novelty mask & $-1.981$ & $[-2.019,-1.945]$ & $[-2.040,-1.922]$ \\
\bottomrule
\end{tabular}

\vspace{0.7em}
\noindent\textit{Panel D: response-witness signed gap from the full-data DFT risk at 20\% (percent).} The bootstrap intervals also appear in main-text Table~\ref{M-tab:response_crossover}; the seed values and paired-$t$ intervals are retained here.\\[2pt]
\resizebox{\textwidth}{!}{%
\begin{tabular}{llrrrrrrr}
\toprule
Dataset & Endpoint & Seed 1 & Seed 2 & Seed 3 & Seed 4 & Seed 5 & Bootstrap 95\% interval & Paired-$t$ 95\% interval \\
\midrule
Carbon & Mean & $+0.167$ & $+0.157$ & $+0.117$ & $+0.154$ & $+0.126$ & $[+0.127,+0.161]$ & $[+0.117,+0.171]$ \\
Carbon & Validation-selected stratum & $+0.073$ & $+0.037$ & $+0.014$ & $+0.057$ & $+0.015$ & $[+0.019,+0.059]$ & $[+0.007,+0.071]$ \\
Carbon & High-force tail & $-0.825$ & $-1.001$ & $-1.312$ & $-0.772$ & $-0.723$ & $[-1.132,-0.763]$ & $[-1.225,-0.629]$ \\
Carbon & High-novelty mask & $-0.239$ & $-0.299$ & $-0.334$ & $-0.227$ & $-0.268$ & $[-0.308,-0.240]$ & $[-0.328,-0.219]$ \\
rMD17 full & Mean & $-0.157$ & $-0.076$ & $-0.139$ & $-0.105$ & $-0.055$ & $[-0.139,-0.073]$ & $[-0.159,-0.054]$ \\
rMD17 full & Validation-selected stratum & $-0.210$ & $-0.123$ & $-0.177$ & $-0.159$ & $-0.130$ & $[-0.188,-0.133]$ & $[-0.204,-0.116]$ \\
rMD17 full & High-force tail & $-0.695$ & $-0.770$ & $-0.774$ & $-0.627$ & $-0.714$ & $[-0.760,-0.669]$ & $[-0.791,-0.641]$ \\
rMD17 full & High-novelty mask & $-0.106$ & $-0.138$ & $-0.176$ & $-0.114$ & $-0.091$ & $[-0.153,-0.102]$ & $[-0.166,-0.084]$ \\
\bottomrule
\end{tabular}%
}
\end{table*}

\subsection{Data and code map}
\label{sec:si_reproducibility_manifest}

The source-data tables are associated with \url{https://doi.org/10.5281/zenodo.22263567}. They contain five-seed aggregate values for the 1\%, 5\% and 20\% budget curves, seed-resolved values for the primary 5\% comparisons, and seed-resolved risks and paired contrasts for the 20\% response crossover. The latter include signed comparisons with the full-data reference. The accompanying archive combines these source tables with the manuscript, data dictionary, analysis scripts and quantitative-figure renderers. Running \path{reproduce.sh} recalculates the paired contrasts, checks them against the source tables and regenerates the quantitative figures without model training.

The paired 5\% witness--FPS/DPP absolute-gap intervals in Supplementary Table~\ref{tab:si_precision_summary} are calculated from \path{data/si/figS5_seed_strata.csv}; the updated output is \path{analysis/generated/tableS7_5pct_witness_vs_coverage.csv}. The other paired statistics are regenerated from the same 5\% seed table and the 20\% response seed table. These derived files are kept separate from the recorded source tables. The companion development repository is \url{https://github.com/JIABI/QUESTS_physical_kernel}. Supplementary Table~\ref{tab:si_repro_manifest} describes the archive scope, and Supplementary Table~\ref{tab:si_figure_source_map} maps figures and tables to their inputs.

\begin{table*}[!tp]
\centering
\small
\caption{\textbf{Source-data and analysis archive scope.} Locations are relative to the archive root. The reproduction entry point checks paired-statistic arithmetic and agreement with the corresponding source tables.}
\label{tab:si_repro_manifest}
\begin{tabular}{p{3.7cm}p{6.2cm}p{6.8cm}}
\toprule
Artifact & Location & Scope \\
\midrule
Main-text source tables & \path{data/main/} & CSV tables for main-text figures and tables, including the 20\% paired response contrast \\
Supplementary source tables & \path{data/si/} & CSV tables including seed-level and budget-sensitivity analyses \\
Method metadata & \path{method_metadata/} & Endpoint-membership and calibration metadata, physical-group definitions and coverage-comparator choices \\
Release metadata & \path{README.md}, \path{DATA_DICTIONARY.md}, \path{VALUE_PROVENANCE.md}, \path{FILE_INVENTORY.csv} & Definitions, recorded and derived values, and file inventory \\
Analysis and tests & \path{analysis/}; \path{reproduce.sh} & Fully enumerated paired-bootstrap and paired-$t$ comparisons; no model training \\
Quantitative-figure code & \path{plotting/} & Portable renderers for Main Figs. 2--5 and the six Supplementary Figures \\
Training configuration & \path{training_configs/} & Shared MACE compression-training settings, provided in the Carbon FPS/DPP 5\%, seed-1 configuration \\
Manuscript and figure assets & \path{manuscript/}; \path{figures/} & Main article, SI, editable LaTeX sources and figure assets \\
\bottomrule
\end{tabular}
\end{table*}

\begin{table*}[!tp]
\centering
\scriptsize
\caption{\textbf{Result-figure and result-table source-data map.} Paths are relative to the aggregate Zenodo release. Wildcards denote all CSV files with the stated prefix. Historical file basenames are retained for compatibility and need not match the current figure or table number; the data dictionary maps legacy fields to the displayed endpoints.}
\label{tab:si_figure_source_map}
\begin{tabular}{p{3.3cm}p{13.5cm}}
\toprule
Item & Archived source-data file(s) \\
\midrule
Main Fig. 1 & Author-supplied conceptual workflow; no numerical source table \\
Main Fig. 2 & \path{data/si/figS6_budget_sensitivity.csv}; \path{data/main/table2_20pct_response_crossover.csv} \\
Main Fig. 3 & \path{data/main/fig5_witness_effects.csv}; \path{data/main/fig5_repair_effects.csv}; auxiliary OBNL quantities in \path{data/main/fig5_obnl_overlap.csv} \\
Main Fig. 4 & \path{data/main/fig4_foundation_summary.csv}; \path{data/main/fig4_finetuning_outcomes.csv}; \path{data/si/figS3_seed_outcomes.csv}; \path{data/si/tableS3_foundation_error.csv} \\
Main Fig. 5 & \path{data/main/fig2_rank_summary.csv}; \path{data/si/figS1_budget_summary.csv}; \path{data/main/fig2_weights.csv}; \path{data/main/fig2_logo_auroc.csv}; \path{data/main/fig2_logo_delta.csv} \\
Main Table 1 & \path{data/si/figS5_seed_strata.csv}; regenerated in \path{analysis/generated/table1_5pct_random_vs_coverage.csv} \\
Main Table 2 & \path{data/main/table2_20pct_response_crossover.csv}; signed-gap intervals from \path{data/si/tableS11_20pct_response_paired_summary.csv}, reproduced in Supplementary Table~\ref{tab:si_20pct_paired}, Panel D \\
Supplementary Fig. S1 & \path{data/si/figS1_budget_summary.csv}; \path{data/main/fig2_weights.csv}; \path{data/si/figS1_logo_auroc.csv}; \path{data/si/figS1_logo_group_delta.csv}; \path{data/si/figS1_logo_delta_summary.csv}; \path{data/si/tableS1_learned_similarity.csv} \\
Supplementary Fig. S2 & \path{data/si/figS3_foundation_summary.csv}; \path{data/main/fig4_finetuning_outcomes.csv}; \path{data/si/figS3_seed_outcomes.csv}; \path{data/si/figS3_paired_differences.csv}; \path{data/si/tableS3_foundation_error.csv} \\
Supplementary Fig. S3 & \path{data/si/figS4_*.csv} \\
Supplementary Fig. S4 & \path{data/si/figS5_seed_strata.csv}; \path{data/si/figS6_budget_sensitivity.csv} \\
Supplementary Fig. S5 & \path{data/si/figS6_budget_sensitivity.csv}; \path{data/main/table4_empirical_resolution_budget.csv} \\
Supplementary Fig. S6 & \path{data/si/tableS9_positive_control.csv}; \path{data/si/figS7_positive_control_summary.csv}; \path{data/si/figS7_positive_control_weights.csv} \\
Supplementary Table S0 & Panel-construction procedure in Supplementary Methods; \path{method_metadata/comparator_validation.csv}; \path{method_metadata/direct_worst_stratum_freeze_v3.csv} \\
Supplementary Table S1 & \path{data/si/tableS1_learned_similarity.csv} \\
Supplementary Table S2 & \path{data/si/tableS2_residual_gap.csv} \\
Supplementary Table S3 & \path{data/si/tableS3_foundation_error.csv} \\
Supplementary Table S4 & \path{data/si/tableS4_response_audit.csv} \\
Supplementary Table S5 & \path{data/si/tableS5_outcome_gap.csv} \\
Supplementary Table S6 & \path{data/si/tableS6_budget_sensitivity.csv} \\
Supplementary Table S7 & Panel A: \path{analysis/generated/tableS7_5pct_witness_vs_coverage.csv}, derived from \path{data/si/figS5_seed_strata.csv}; Panel B: \path{data/si/tableS7_empirical_resolution_materiality.csv}; Panel C: \path{analysis/generated/table1_5pct_random_vs_coverage.csv}. \\
Supplementary Table S8 & \path{data/si/tableS8_gamma_stability.csv} \\
Supplementary Table S9 & \path{data/si/tableS9_positive_control.csv} \\
Supplementary Table S10 & \path{data/si/figS5_seed_strata.csv}; \path{data/si/figS6_budget_sensitivity.csv}; grouped intervals in the files mapped above \\
Supplementary Table S11 & \path{data/si/figS5_seed_strata.csv}; \path{data/main/table2_20pct_response_crossover.csv}; \path{data/si/tableS11_20pct_response_seed_results.csv}; \path{data/si/tableS11_20pct_response_paired_summary.csv} \\
\bottomrule
\end{tabular}
\end{table*}

\begin{samepage}
\subsection{Generative-AI assistance during preparation}
\label{sec:si_ai_assistance}
AI-assisted tools were used during
manuscript preparation for language editing, proofreading and quality-control
support. They did not generate or modify primary data, perform the reported
numerical analyses, select models or decision rules, or create images or
multimedia. The authors verified all AI-assisted output and take responsibility
for the submitted work.
\end{samepage}

\clearpage
\raggedbottom
\bibliography{rsc}
\bibliographystyle{rsc}